\documentclass[12pt]{article}

\usepackage{amsmath}
\usepackage{amsfonts}
\usepackage{graphicx,psfrag,epsf}
\usepackage{caption}
\usepackage{subcaption}
\usepackage{enumerate}
\usepackage{bm}
\usepackage{listings}
\usepackage{mathtools}

\usepackage[colorlinks=true,allcolors=blue]{hyperref}

\usepackage[style=authoryear,backref=true]{biblatex}
\bibliography{refs}
\bibliography{zotero}

\newcommand{\blind}{0}

\newcommand{\mytitle}{Multi-modal Bayesian Neural Network Surrogates with Conjugate Last-Layer Estimation}

\begin{document}

\def\spacingset#1{\renewcommand{\baselinestretch}%
{#1}\small\normalsize} \spacingset{1}

%%%%%%%%%%%%%%%%%%%%%%%%%%%%%%%%%%%%%%%%%%%%%%%%%%%%%%%%%%%%%%%%%%%%%%%%%%%%%%

\if0\blind
{
  \title{\bf \mytitle}
  \author{Ian Taylor\footnote{Corresponding Author: ian.taylor@nlr.gov; ian.m.taylor1@gmail.com}\\
    National Laboratory of the Rockies\\
    and \\
    Juliane Mueller \\
    National Laboratory of the Rockies\\
    and \\
    Julie Bessac \\
    National Laboratory of the Rockies}
  \maketitle
} \fi

\if1\blind
{
  \bigskip
  \bigskip
  \bigskip
  \begin{center}
    {\LARGE\bf \mytitle}
\end{center}
  \medskip
} \fi

\bigskip
\begin{abstract}
As data collection and simulation capabilities advance, multi-modal learning, the task of learning from multiple modalities and sources of data, is becoming an increasingly important area of research. Surrogate models that learn from data of multiple auxiliary modalities to support the modeling of a highly  expensive quantity of interest have the potential to aid  outer loop applications such as  optimization, inverse problems, or sensitivity analyses when multi-modal data are available. We develop two multi-modal Bayesian neural network surrogate models and leverage conditionally conjugate distributions in the last layer to estimate model parameters using stochastic variational inference (SVI). We provide a method to perform this conjugate SVI estimation in the presence of partially missing observations. We demonstrate improved prediction accuracy and uncertainty quantification compared to uni-modal surrogate models for both scalar and time series data.
\end{abstract}

\noindent%
{\it Keywords:}  Bayesian neural networks, multi-modal learning,  surrogate models, variational inference

\newpage

%\todo{Overall - standardize between ``objective function'' and ``quantity of interest''. Probably ``quantity of interest'', since reducing the focus on BO}

\spacingset{1.45}
\section{Introduction and Background} \label{sec:introduction}
%\julie{i think that the emphasize on BO is too strong here compared to the already existing chalenges of multimodalities. also if we focus so much on BO, we have to later demonstrate why these models are better than others. }
%Multi-modal data--data from multiple sources and taking multiple forms--are increasingly common in science, machine learning, and artificial intelligence. With this data comes the desire to use it to make scientific discovery faster and more efficient, especially when expensive computational resources are required. We aim to incorporate multi-modal data into surrogate models that can be leveraged to  optimize functions calculated by black-box computer simulations.
%\julie{what are the ML/stats challenges of working with multimodalities? this is important to present right away. }

%\julie{I adjusted a bit that paragraph and i think it is nice as it is now}
Many computational methodologies and applications, such as expensive black-box optimization, are often tackled by leveraging surrogate models that map inputs to outputs and guide an adaptive search for improved solutions. Advances in simulation as well as experimentation capabilities have allowed us to collect vastly more data and data of different types (e.g., images, time series, and text) than can be handled by widely used surrogate models such as Gaussian processes (GPs). To maximize the usefulness of collected data, advances in multi-modal surrogate modeling are needed as well as leveraging dependencies between modalities. Ideally, such multi-modal surrogates provide us with predictions of the quantities of interest as well as with uncertainty estimates of these predictions. 
%with the goal to exploit this information in iterative data acquisition decisions, as is done in Bayesian optimization (\todo{citations}). 

%\julie{Let's shrink this and cast it as a motivating application of multi-modal surrogates}
In this work, we design two multi-modal surrogates following the motivating example of Bayesian optimization (BO). BO is a gradient-free optimization method for expensive black-box objective functions \parencite{jones_efficient_1998}. The BO procedure begins with an initial design of input values to span the search space, from which a surrogate model is trained to approximate the objective function. Then, an acquisition function guides observation of the objective function at new points and the surrogate model is updated to include each additional observation. In this paper, we focus on the surrogate model for the objective function, which approximates the objective function at each step based on the expensive evaluations performed so far. Most commonly, the surrogate model is a Gaussian process that allows for statistically principled modeling and uncertainty quantification. However, other surrogate models have been explored, including neural networks \parencite[see][]{li_study_2023, foldager_role_2023, brunzema_bayesian_2024} and random forests \parencite[e.g.,][]{mccullough_high-throughput_2020, williams_enabling_2020, jayarathna_experimental_2024}. Uncertainty quantification in the surrogate model is important for the BO task, as the acquisition function, which determines the sequential selection of new points to  evaluate, seeks to  balance exploration and exploitation \parencite{jones_efficient_1998, jones_taxonomy_2001, de_ath_greed_2021}. Because surrogate models have broader applications than BO, we will refer to the modeled function as a quantity of interest and only use the term ``objective function'' when specifically referencing the motivating application of optimization. %Acquisition function such as expected improvement (\todo{cite}), upper/lower confidence bound (\todo{cite}) and probability of improvement (\todo{cite}) combine the expected value of the surrogate model and its estimated uncertainty at a point to find the next best candidate. 

In this paper, we introduce two novel multi-modal Bayesian neural network (BNN) surrogate models motivated by BO for multi-modal data. The remainder of this section contains necessary background for the methods used. Section \ref{sec:multi-modal-surrogate-models} introduces the two proposed surrogate models. We leverage conditionally conjugate distributions in the network's last layer in stochastic variational inference (SVI) and provide a method to perform this conjugate SVI estimation in the presence of partially missing observations (Section \ref{sec:estimation}). We demonstrate improved prediction accuracy and uncertainty quantification compared to unimodal surrogate models for both scalar and time series data (Section \ref{sec:simulation}), and examine the performance of multi-modal models on real-world data (Section \ref{sec:application}). Finally, we summarize our results and propose future  research directions (Section \ref{sec:conclusion}).

%\julie{awesome!}

\subsection{Multi-modal Learning}

Multi-modal learning is the area of machine learning concerned with combining multiple kinds, or modalities, of data to accomplish a common task. Multimedia search \parencite[e.g.,][]{lan_multimedia_2014}, image generation \parencite[e.g.,][]{koh_generating_2023}, and speech synthesis \parencite[e.g.,][]{hunt_unit_1996, ma_unpaired_2019} are all examples of multi-modal machine learning tasks involving two or more of the image, audio, video, and text modalities. Multi-modal learning poses unique challenges due to the heterogeneity of the data, which can be placed into one of five categories \parencite{baltrusaitis_multimodal_2019}:
\begin{enumerate}
    \item Multi-modal representation: the task of summarizing multi-modal data to capture information common to each modality,
    \item Multi-modal translation: the task of converting data of one modality into a different modality,
    \item Multi-modal alignment: identifying direct relationships between components of data of multiple modalities,
    \item Multi-modal fusion: integrating information from multiple modalities with the goal of predicting some outcome measure,
    \item Multi-modal co-learning: transferring knowledge between modalities.
\end{enumerate} 
The challenge within multi-modal learning that is most relevant to our goal of a multi-modal surrogate model is multi-modal fusion from the above list. Returning to our motivating example, multi-modal surrogate models for BO are a multi-modal fusion problem in the sense that they incorporate data from multiple sources and multiple modalities to more accurately predict unobserved values of the quantity of interest. In this work, we consider the special case of multiple data sources of the same nature (for instance all sources are time series), which is a common multi-modal setting in scientific data. We still refer to our modeling strategy as multi-modal. 

The earliest work in multi-modal fusion was in speech recognition from audio and visual recordings \parencite{yuhas_integration_1989}. Since then, multi-modal fusion has been applied to cardiovascular disease diagnosis \parencite[e.g.,][]{yoon_multi-modal_2023}, object classification \parencite[e.g.,][]{gehler_feature_2009, bucak_multiple_2014}, and emotion recognition \parencite[e.g.,][]{castellano_emotion_2008, wollmer_context-sensitive_2010, chen_multi-modal_2015}, among other fields. Due to this long and diverse history, a variety of methods have been proposed and used for multi-modal fusion, including model-agnostic (early, late, and hybrid fusion) and model-based methods. Despite this history, multi-modal fusion can still struggle to perform well when there are varying levels of noise in the modalities, or to capture complementary information in the modalities \parencite{baltrusaitis_multimodal_2019}.

In contrast to multi-modal fusion applications like speech recognition, or audio/visual classification in which complex or high-dimensional modalities are used as predictors for a relatively simple outcome, multi-modal surrogate models map a simpler input domain to potentially complex modalities. In other words, the multi-modal data itself may be the outcome measure of interest. We propose a model-based neural network approach to construct our surrogate models, paired with dimension reduction of the more complex modalities, to solve this problem.

\subsection{Multi-fidelity Surrogate Models} \label{sec:background-multi-fidelity} 

Multi-fidelity surrogate models belong to the class of multi-modal surrogate models and incorporate lower-fidelity, less computationally expensive views of the quantity of interest in order to more quickly and cheaply estimate the quantity of interest. One particular example of multi-fidelity applications are so-called multi-level simulations, in which the resolution can be adjusted to trade-off speed and accuracy. For example, if the quantity of interest is calculated via an expensive high-resolution fluid simulation, then the same simulation may be run at a lower resolution to quickly provide an approximate, potentially biased or noisy version of the objective quantity.

%\julie{could we say that co-kriging can be used in the more general context of multi-modality under linear assumptions? I'd like to shrink that section to the benefit of the multi-modal one. we need to lighten the references to objective functions as we can a lot already without them. }
Co-Kriging \parencite{kennedy_predicting_2000} uses multiple Gaussian processes to simultaneously model low- and high-fidelity objective functions through a linear relationship. The linear relationship follows from an assumption of limited dependence between the functions, specifically, if $z_t(\bm{x})$ and $z_{t-1}(\bm{x})$ are models of the functions with fidelity $t$ and the next-highest fidelity, $t-1$, respectively, the co-Kriging model assumes $\mathrm{Cov}\{z_t(\bm{x}),z_{t-1}(\bm{x}')|z_{t-1}(\bm{x})\} = 0$, for $\bm{x} \neq \bm{x}'$. The resulting model is
\begin{equation}
    z_t(\bm{x}) = \rho_{t-1}z_{t-1}(\bm{x}) + \delta_t(\bm{x}),
\end{equation}
where $\rho_{t-1}$ is a constant and $\delta_t(\bm{x})$ is an independent stationary Gaussian process. Hierarchical Kriging \parencite{han_hierarchical_2012} is an alternative multi-fidelity surrogate model based on Gaussian processes. By contrast to co-Kriging, hierarchical Kriging uses only the expected value of the lower-fidelity surrogate models in its linear trend, which simplifies the process of fitting the model, and in multi-fidelity BO, allows for more flexibility in selecting the acquisition function used later. Recent work in multi-fidelity surrogate models has branched out from the more conventional Gaussian process-based models into models from the machine learning literature, for example, neural network-based multi-fidelity surrogate models have been developed \parencite{li_multi-fidelity_2020, zhang_multi-fidelity_2021}. Co-Kriging and hierarchical Kriging can be used directly in the more general context of multi-modal data if all observations or functions have scalar values and the relationships are assumed to be linear.

In contrast to multi-fidelity data, the relationships between modalities of multi-modal data are more complex, potentially non-linear or discontinuous. Therefore, existing multi-fidelity surrogate models that assume a linear relationship between the data sources such as co-Kriging and hierarchical Kriging may not capture these relationships if applied directly to multi-modal data, and therefore may not offer any improvement in prediction over uni-modal surrogate models. We seek to fill the need for surrogate models that can leverage these more complex relationships to improve prediction.

\subsection{Bayesian Neural Networks} \label{sec:bayesian-neural-networks}

Bayesian Neural Networks (BNNs) are neural networks where the parameters (weights and biases) are given prior distributions and the posterior distributions of the parameters are estimated on observed data. In this paper, we focus on BNN surrogate models due to their use in multi-modal fusion problems, their flexibility, and their relationship to more conventional Gaussian process surrogate models in BO. \textcite{bickel_priors_1996} show that an infinitely-wide BNN with a single hidden layer and Gaussian priors is a Gaussian process with a neural covariance function. \textcite{lee_deep_2018} later show that deep infinitely-wide BNNs are also Gaussian processes. However, by using finite BNNs instead of GPs or infinite BNNs, we avoid assuming stationarity of variance in the model or needing to calculate the neural covariance function. The use of a BNN also allows us to produce a posterior sample of the weights of the network, amortizing the cost of future evaluations of the surrogate model at unobserved locations.

We consider surrogate models based on fully-connected BNNs of the form
\begin{align}
    \bm{z}_0 &\coloneq \bm{x} \label{eqn:neural-net-start} \\
    \bm{z}_k &\coloneq \sigma\left(s_{k}\cdot\left(\frac{1}{\sqrt{h_{k-1}}}\bm{W}_{k-1}\bm{z}_{k-1} + \bm{b}_{k-1}\right)\right),\quad k=1,\dots,\ell \label{eqn:neural-net-layer}\\
    NN(\bm{x}; \bm{W}_0,\dots,\bm{W}_{\ell},\bm{b}_0,\dots,\bm{b}_{\ell}) &\coloneq  \frac{1}{\sqrt{h_\ell}}\bm{W}_\ell \bm{z}_\ell + \bm{b}_\ell, \label{eqn:neural-net-end}
\end{align}
where the parameters $\bm{W}_k$ and $\bm{b}_k$ for $k=0,\dots,\ell$ are the weights and biases, respectively, of the $k^\mathrm{th}$ layer, $h_k$ is the dimension of the vector $\bm{z}_k$, $\sigma(\cdot)$ is a non-linear activation function applied element-wise, e.g., ReLU or Tanh, and $s_k > 0$ is a scale factor adjusting the slope of the activation function in each layer. The values $\bm{z}_k$ for $k=1,\dots,\ell$ are the activation values of the hidden layers in the network, and $NN$ is the final neural network.

We place independent standard Gaussian ($\mathrm{N}(0,1)$) priors on each weight and bias in the network, and independent $\mathrm{Gamma}(2,1)$ priors on the scale parameters, $s_k$. The priors combine with the factor of $h_k^{-1/2}$ in each layer to ensure a finite prior variance of the layer activations $\bm{z}_k$. The factor of $h_k^{-1/2}$ also appears in \textcite{bickel_priors_1996} and \textcite{lee_deep_2018} where it is important to insure the convergence to a Gaussian process as $h_k \to \infty$. By including it here, we are using neural networks that can be thought of as truncations of these limits, thus approximating Gaussian processes.

We assemble these networks into composite models to model observations $\{(\bm{x}_i,\bm{y}_i)\}_{i=1}^n$ of a main data modality, and auxiliary observations $\{(\bm{x}^{(m)}_j,\bm{y}^{(m)}_j)\}_{j=1}^{n_m}$ for $m=1,\dots,M$ from $M$ additional data modalities. We outline two main architectures in Section \ref{sec:multi-modal-surrogate-models}, a joint model and a layered model.

%Bayesian neural networks, like all Bayesian models, are estimated through the posterior distribution of their parameters. If the parameters are estimated with posterior mean point estimates, then the estimates produced are equivalent to those of a deterministic neural network with regularization on the weights. So to fully realize the benefits of the Bayesian approach, we estimate the posterior distribution as a whole. 

BNN posteriors can be estimated either by an ensemble of samples, e.g., Markov chain Monte Carlo \parencite{bickel_monte_1996} and Deep Ensembles \parencite{lakshminarayanan_simple_2017}; or by approximate closed-form distributions, e.g., MC-Dropout \parencite{gal_dropout_2016}, Laplace approximations \parencite{mackay_practical_1992},  Stochastic Weight Averaging Gaussian \parencite{maddox_simple_2019}, and stochastic variational inference \parencite[SVI;][]{blundell_weight_2015}. MCMC estimates the posterior by producing auto-correlated samples that (as the number of samples approaches infinity) follow the desired posterior distribution. Despite being very commonly used for MCMC on BNN models, for large networks, the No U-Turn Sampler \parencite[NUTS,][]{hoffman_no-u-turn_2014} struggles both in terms of speed and exploration of the posterior. Deep Ensembles \parencite{lakshminarayanan_simple_2017} use multiple initialization points and optionally multiple bootstrap samples of data to produce a random ensemble of trained networks. With bootstrap samples of data, this ensemble can approximate the posterior distribution for uninformative priors.

MC-Dropout estimates uncertainty by using Dropout \parencite{srivastava_dropout_2014} at inference time in addition to training time. MC-Dropout thus approximates the posterior distributions of weights as scaled Bernoulli distributions. Laplace approximations use curvature at the loss function minimum to approximate the posterior as a Gaussian distribution. Stochastic Weight Averaging Gaussian also approximates the posterior as a Gaussian distribution, but uses points from the end of the standard training process using stochastic gradient descent instead of the loss function's curvature to calculate the first and second moments.

We focus on SVI for our task to take advantage of the network structure in specifying the approximate posterior distribution. SVI approximates the posterior of the network parameters by finding the closest match from an analytical family of distributions. For a model with parameters, $\bm{\theta}$, and data, $\bm{x}$, we want to find a closed-form distribution $Q(\bm{\theta})$, which minimizes the KL-divergence, $D_{KL}(Q\|p)$, from the true posterior distribution, $p(\bm{\theta}|\bm{x})$. 
Commonly, the family for $Q$ is chosen to be the so-called mean-field approximation: independent Gaussian distributions over each component of $\bm{\theta}$ (possibly after first transforming constrained parameters into unconstrained space). This choice makes the problem of optimizing $Q$ computationally tractable \parencite{zhang_theoretical_2020, sheng_stability_2025}, and has been applied to estimate posterior distributions in applications such as genomics \parencite[e.g.,][]{carbonetto_scalable_2012} and language modeling \parencite[e.g.,][]{blei_latent_2003}. However, for posterior distributions with strong dependence between parameters, the approximation can be poor \parencite[e.g.,][]{wang_lack_2004}. BNN posterior distributions do have strong dependence between parameters.

Bayesian last-layer neural networks \parencite[BLL;][]{lazaro-gredilla_marginalized_2010} are also commonly used \parencite[e.g.,][]{snoek_scalable_2015}. For BLLs, weights and biases in all layers except the last layer are estimated with point estimates, while weights and biases in the last layer are treated as random variables. As a consequence, there is less variability in the estimation of a BLL than a fully Bayesian neural network, where all weights and biases are treated as random variables. The distribution of the weights and biases in the last layer is often comprised of independent Gaussian distributions; however, distributions with more complex dependence have been proposed \parencite[e.g.,][]{li_multi-fidelity_2020, harrison_variational_2024}. The estimation method we propose in Section \ref{sec:estimation} incorporates dependence through conjugate full conditional distributions in the last layer.

\section{Proposed Multi-modal Surrogate Models} \label{sec:multi-modal-surrogate-models}

We propose two BNN-based multi-modal surrogate models: a joint model and a layered model. The models are influenced by different surrogate modeling and multi-modal learning methods, and are distinct in their use of data from different modalities. The joint model models values from all data sources as one combined vector output from a single BNN. The layered model is comprised of two BNNs, one modeling auxiliary sources and one modeling the main quantity of interest. Output from the network modeling the auxiliary sources, including predicted values between observations, is used as additional input to the main network. These models are trained on observations $\{(\bm{x}_i,\bm{y}_i)\}_{i=1}^n$ of a main data modality, and auxiliary observations $\{(\bm{x}^{(m)}_j,\bm{y}^{(m)}_j)\}_{j=1}^{n_m}$ for $m=1,\dots,M$ from $M$ additional data modalities.

\subsection{Joint Model} \label{sec:model-joint}

\begin{figure}
    \centering
    \begin{subfigure}{0.5\textwidth}
        \includegraphics[width=\linewidth]{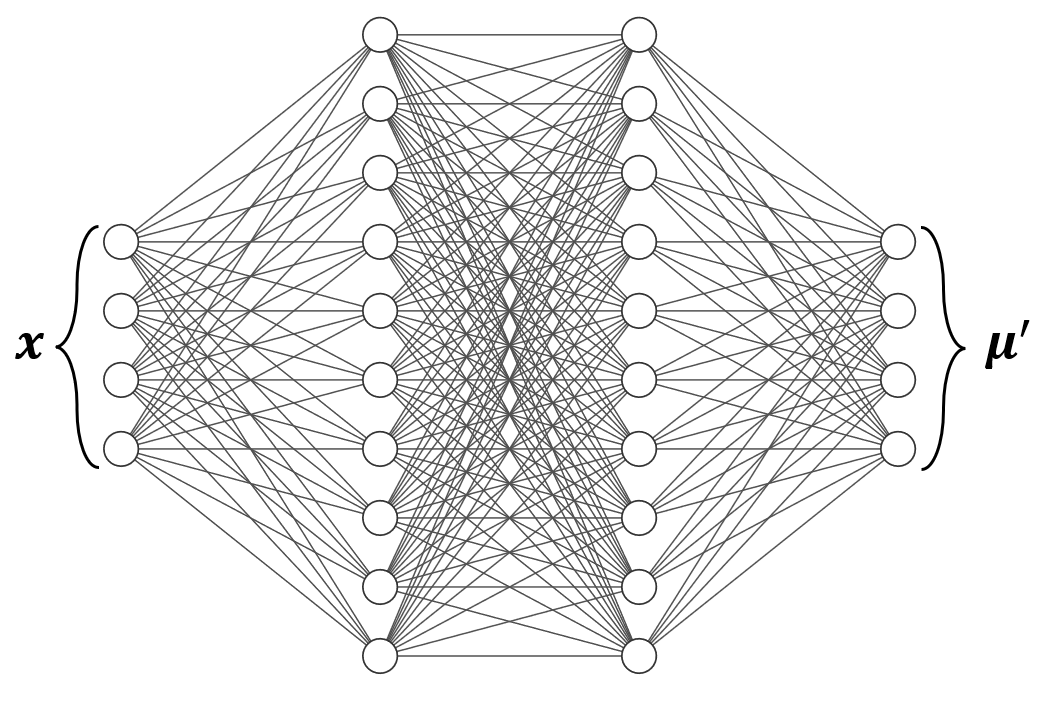}
        \subcaption{Joint model}
        \label{fig:architecture-joint-example}
    \end{subfigure}%
    \begin{subfigure}{0.5\textwidth}
        \includegraphics[width=\linewidth]{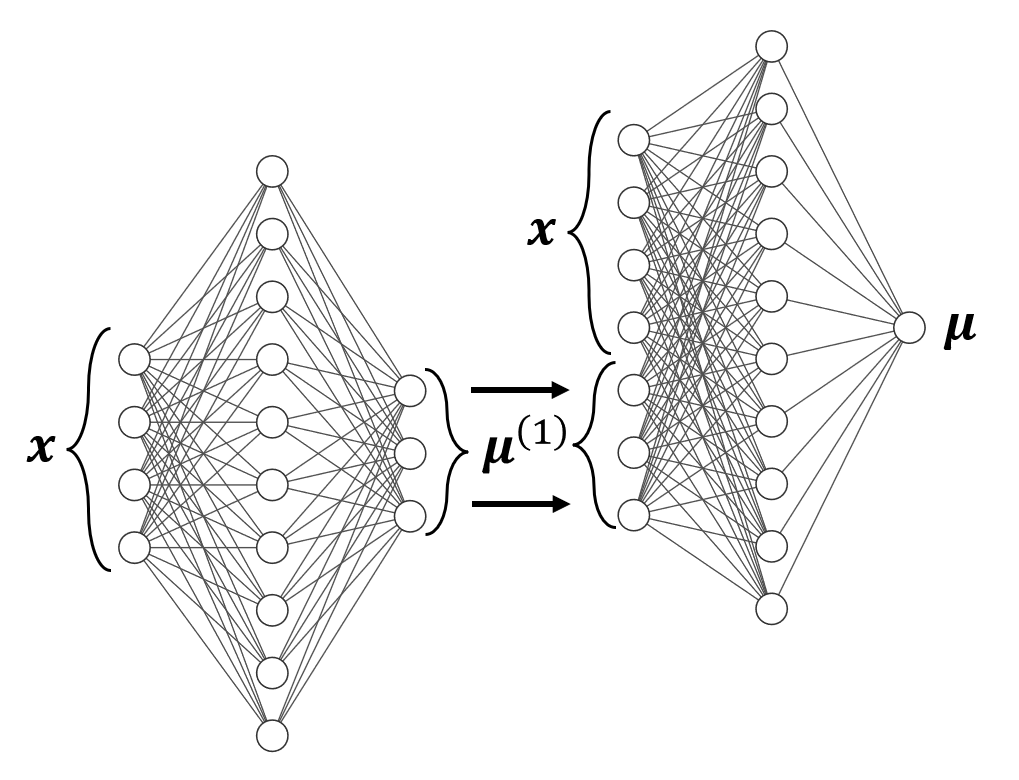}
        \subcaption{Layered model}
        \label{fig:architecture-layered-example}
    \end{subfigure}
    \caption{Examples of the joint (left) and layered (right) model architectures for data with a scalar quantity of interest and one 3-dimensional vector auxiliary modality. In the joint model, the output, $\bm{\mu}'$ models the mean for both the main response, $y$, and the auxiliary response, $\bm{y}^{(1)}$, as a single vector $\bm{\mu}'=(\mu,\ \bm{\mu}^{(1)\top})^\top$. In the layered model, the mean of the auxiliary response, $\bm{\mu}^{(1)}$, is modeled by a separate BNN surrogate model and is then used as input to the main BNN surrogate model.}
    \label{fig:architecture-examples}
\end{figure}

The most straightforward way to model multiple data modalities is as the joint output of a single BNN. In other words, we construct a model,
\begin{align}
    \bm{y}' &\sim \mathrm{N}(\bm{\mu}', \bm{\Sigma}) \label{eqn:augmented-response-variable}\\
    \bm{\mu}' &= NN(\bm{x}; \bm{W}_0,\dots,\bm{W}_{\ell},\bm{b}_0,\dots,\bm{b}_{\ell}) \\
    {\bm{\Sigma}}^{-1} &\sim \mathrm{Wishart}(\nu_0, \bm{V}_0), \label{eqn:joint-model-priors-start}\\
    \begin{bmatrix}\bm{b}_\ell & \bm{W}_\ell\end{bmatrix} | \bm{\Sigma} &\sim \mathrm{MatrixNormal}(\bm{0}, \bm{\Lambda}_0, \bm{\Sigma})\\
    \bm{W}_k &\sim \mathrm{N}(0,1),\quad k=0,\dots,\ell-1 \\
    \bm{b}_k &\sim \mathrm{N}(0,1),\quad k=0,\dots,\ell-1 \label{eqn:joint-model-end}
\end{align}
where $\bm{y}' = \begin{bmatrix}\bm{y}^\top&\bm{y}^{(1)\top}&\dots&\bm{y}^{(M)\top}\end{bmatrix}^\top$ is an output vector comprised of all modalities concatenated together, and $NN(\cdot;\cdot)$ is a BNN as defined in \eqref{eqn:neural-net-start}-\eqref{eqn:neural-net-end}. We call this model the ``joint model," and show a simplified illustration of this model in Figure \ref{fig:architecture-joint-example}. The statistical parameters of this model are $\bm{\Sigma}$ and $\bm{b}_k, \bm{W}_k$ for $k=0,\dots,\ell$. The model is estimated by estimating the posterior distributions of these parameters with closed form variational approximations. We expand on the specific choice of priors \eqref{eqn:joint-model-priors-start}-\eqref{eqn:joint-model-end} and their significance to estimation in Section \ref{sec:estimation}. Intuitively, we expect that this kind of model would learn representations of the input, $\bm{x}$, in its hidden layers that capture shared qualities of all data modalities simultaneously.

Multi-modal representation learning has been a core component of multi-modal learning for a long time, specifically when using a common representation for downstream tasks. \textcite{ngiam_multimodal_2011} use bimodal deep autoencoders to place audio and visual data of people speaking into a shared representation space, before using that shared representation to classify the spoken syllables. The representation mapping of the input has also been used in multi-fidelity BO to optimize more challenging functions \parencite{raissi_deep_2016}, allowing for more flexible covariance between observations that can model functions with discontinuities that would be difficult for conventional Gaussian processes. While this method combines a deterministic neural network representation of the input space with a GP correlation structure, our proposed method uses a BNN for both the representation (hidden layers) and correlation structure between modalities (last layer) of the model.

%In the motivating application of BO, not every modality will be observed for the same input values as the acquisition function adaptively guides sampling of each modality. This results in joint observations, $\bm{y}'$, with missing values. Therefore this model must be  trainable in the presence of partially missing response values. We discuss our approach to handling partially missing observations in Section \ref{sec:estimation}.

\subsection{Layered Model} \label{sec:model-layered}

The layered model is an alternate approach that draws more from the structure of surrogate models for multi-fidelity BO, co-Kriging and hierarchical Kriging. In this model, the additional data modalities are used as predictors in a neural network surrogate model for the main modality. Because there may not be observations of the additional modalities at the same input values as the observations of the main modality, we use separate neural network surrogate models to predict unobserved values of the additional modalities with uncertainty. This results in a model comprised of $M$ auxilliary networks, for $m=1,\dots,M$:
\begin{align}
    \bm{y}^{(m)} &\sim \mathrm{N}(\bm{\mu}^{(m)},\bm{\Sigma}^{(m)}), \\
    \bm{\mu}^{(m)} &= NN(\bm{x}; \bm{W}_0^{(m)},\dots,\bm{W}_{\ell}^{(m)},\bm{b}_0^{(m)},\dots,\bm{b}_{\ell}^{(m)}), \\
    {\bm{\Sigma}^{(m)}}^{-1} &\sim \mathrm{Wishart}(\nu_0^{(m)}, \bm{V}_0^{(m)}), \label{eqn:layered-model-aux-priors-start}\\
    \begin{bmatrix}\bm{b}_\ell^{(m)} & \bm{W}_\ell^{(m)}\end{bmatrix} | \bm{\Sigma}^{(m)} &\sim \mathrm{MatrixNormal}(\bm{0}, \bm{\Lambda}_0^{(m)}, \bm{\Sigma}^{(m)})\\
    \bm{W}_k^{(m)} &\sim \mathrm{N}(0,1),\quad k=0,\dots,\ell-1 \\
    \bm{b}_k^{(m)} &\sim \mathrm{N}(0,1),\quad k=0,\dots,\ell-1, \label{eqn:layered-model-aux-end}
\end{align}
whose outputs feed into a main network,
\begin{align}
    \bm{y} &\sim \mathrm{N}(\bm{\mu}, \bm{\Sigma}) \\
    \bm{\mu} &= NN(\bm{x},\bm{\mu}^{(1)},\dots,\bm{\mu}^{(M)}; \bm{W}_0,\dots,\bm{W}_{\ell},\bm{b}_0,\dots,\bm{b}_{\ell}) \\
    {\bm{\Sigma}}^{-1} &\sim \mathrm{Wishart}(\nu_0, \bm{V}_0), \label{eqn:layered-model-main-priors-start}\\
    \begin{bmatrix}\bm{b}_\ell & \bm{W}_\ell\end{bmatrix} | \bm{\Sigma} &\sim \mathrm{MatrixNormal}(\bm{0}, \bm{\Lambda}_0, \bm{\Sigma})\\
    \bm{W}_k &\sim \mathrm{N}(0,1),\quad k=0,\dots,\ell-1 \\
    \bm{b}_k &\sim \mathrm{N}(0,1),\quad k=0,\dots,\ell-1. \label{eqn:layered-model-main-end}
\end{align}
We show a simplified example of this model in Figure \ref{fig:architecture-layered-example}. The statistical parameters of this model are $\bm{\Sigma}$, $\bm{b}_k, \bm{W}_k$ for $k=0,\dots,\ell$, $\bm{\Sigma}^{(m)}$ for $m=1,\dots,M$, and $\bm{b}_k^{(m)}, \bm{W}_k^{(m)}$ for $m=1,\dots,M$ and $k=0,\dots,\ell$. The model is estimated by estimating the posterior distributions of these parameters with closed form variational approximations. We expand on the specific choice of priors \eqref{eqn:layered-model-aux-priors-start}-\eqref{eqn:layered-model-aux-end} and \eqref{eqn:layered-model-main-priors-start}-\eqref{eqn:layered-model-main-end} and their significance to estimation in Section \ref{sec:estimation}.

Co-Kriging \parencite{kennedy_predicting_2000} and hierearchical Kriging \parencite{han_hierarchical_2012} use lower-fidelity computer simulations similarly to how this layered model uses alternate modalities. Both models use Gaussian processes to model lower-fidelity computer simulations and use a linear function of these Gaussian processes as the mean of the Gaussian process that models the main quantity of interest. For co-Kriging, the actual lower fidelity GP is used, while in hierarchical Kriging, only the GP's expected value is used. Thus, they both model the objective function as a linear model with lower-fidelity computer simulations as predictors and a GP modeling the residuals. Because our layered model uses output from BNNs modeling alternate modalities as inputs to the BNN modeling the main modality, it is conceptually similar to both of these multi-fidelity BO models. It is more flexible, though, because the BNN allows for non-linear relationships between modalities.

\section{Variational Estimation of Proposed BNN Models} \label{sec:estimation}

In this section, we describe a variational approximation to the posterior of BNN parameters that we use for the joint and layered multi-modal surrogate models. We consider the approximation for a single BNN, then describe the approximation in the presence of missing data. Because each of our proposed models is composed of one or more of these BNNs, the models can be fit by simultaneously estimating all of their component BNNs. The multi-modal surrogate models and conditional last layer estimation are implemented using Pyro \parencite{bingham_pyro_2018, phan_composable_2019}, and are available in the Python package \texttt{mmbo}.\footnote{Code will be made publicly available prior to publication.}

\subsection{Conjugate Last Layer Variational Estimation} \label{sec:estimation-conjugate-last-layers}

Consider a BNN for regression as defined in \eqref{eqn:neural-net-start}-\eqref{eqn:neural-net-end}, with $m$ input dimensions and $k$ output dimensions. Let all weights, biases, and other parameters prior to the last layer be called $\bm{\Phi}$, and the activations of the last layer of $h$ hidden nodes be $\bm{z}_\ell(\bm{x};\bm{\Phi}) \in \mathbb{R}^h$ for input $\bm{x} \in \mathbb{R}^m$, as shown in in \eqref{eqn:neural-net-layer}. Let the weights in the last layer be $\bm{W} \in \mathbb{R}^{h\times k}$, and the biases of the outputs be $\bm{b} \in \mathbb{R}^k$. Then, overall, the network can be expressed as
\begin{equation}
    NN(\bm{x}; \bm{\Phi},\bm{W},\bm{b})^\top \coloneq \bm{b}^\top + \bm{z}_\ell(\bm{x};\bm{\Phi})^\top\bm{W},
\end{equation}
with $NN(\bm{x}; \bm{\Phi},\bm{W},\bm{b}) \in \mathbb{R}^k$. If we then have observations, $\mathcal{D} = \{(\bm{x}_i, \bm{y}_i)\}_{i=1}^n$, we can model them as
\begin{align}
    \bm{y}_i &= NN(\bm{x}_i; \bm{\Phi},\bm{W},\bm{b}) + \bm{\epsilon}_i, \\
    &= \bm{b} + h^{-1/2}\bm{W}^\top \bm{z}_\ell(\bm{x}_i;\bm{\Phi}) + \bm{\epsilon}_i, \quad\text{for $i=1,\dots,n$}, \\
    \bm{\epsilon}_i &\overset{iid}{\sim} N(\bm{0},\bm{\Sigma}), \quad\text{for $i=1,\dots,n$}, \\
    \bm{\Phi} &\sim p(\bm{\Phi}),\  
    \bm{W} \sim p(\bm{W}),\ 
    \bm{b} \sim p(\bm{b}),\ 
    \bm{\Sigma} \sim p(\bm{\Sigma}).
\end{align}

Conditioned on $\bm{\Phi}$, this is a Bayesian multivariate linear regression problem. With conjugate priors, $p(\bm{W})$, $p(\bm{b})$, and $p(\bm{\Sigma})$, the full conditional distribution,
\begin{equation}
    p(\bm{W}, \bm{b}, \bm{\Sigma} | \bm{\Phi}, \mathcal{D}), \label{eqn:last-layer-full-conditional-general}
\end{equation}
has a known analytical form. For the BNN model, the conjugate priors are 
\begin{align}
    \bm{\Sigma}^{-1} &\sim \mathrm{Wishart}(\nu_0, \bm{V}_0) \label{eqn:bnn-priors-Sigma} \\
    \begin{bmatrix}\bm{b} & \bm{W}\end{bmatrix} | \bm{\Sigma} &\sim \mathrm{MatrixNormal}(\bm{0}, \bm{\Lambda}_0, \bm{\Sigma}). \label{eqn:bnn-priors-weights}
\end{align}
With
\begin{equation}
    \bm{Z} = \begin{bmatrix}
        1 & h^{-1/2}\bm{z}_\ell(\bm{x}_1;\bm{\Phi})^\top \\
        \vdots & \vdots \\
        1 & h^{-1/2}\bm{z}_\ell(\bm{x}_n;\bm{\Phi})^\top
    \end{bmatrix}, \quad
    \bm{Y} = \begin{bmatrix}
        \bm{y}_1^\top \\
        \vdots \\
        \bm{y}_n^\top
    \end{bmatrix},
\end{equation}
the full conditional distributions are
\begin{align}
    \bm{\Sigma}^{-1}|\bm{Z},\bm{Y} &\sim \mathrm{Wishart}(\nu_n, \bm{V}_n), \label{eqn:sigma-full-conditional}\\
    \begin{bmatrix}\bm{b} & \bm{W}\end{bmatrix} | \bm{\Sigma},\bm{Z},\bm{Y} &\sim \mathrm{MatrixNormal}(\widehat{\bm{W}}_n, \bm{\Lambda}_n, \bm{\Sigma}), \label{eqn:weight-full-conditional}
\end{align}
where
\begin{align}
    \nu_n &= \nu_0 + n, \label{eqn:full-conditional-params-start}\\
    \bm{V}_n &= (\bm{V}_0^{-1} + (\bm{Y} - \bm{Z}\widehat{\bm{W}}_n)^\top (\bm{Y} - \bm{Z}\widehat{\bm{W}}_n) + \widehat{\bm{W}}_n^\top \bm{\Lambda}_0^{-1} \widehat{\bm{W}}_n)^{-1}, \\
    \widehat{\bm{W}}_n &= \bm{\Lambda}_n \bm{Z}^\top \bm{Y},\\
    \bm{\Lambda}_n &= (\bm{\Lambda}_0^{-1} + \bm{Z}^\top \bm{Z})^{-1}. \label{eqn:full-conditional-params-end}
\end{align}

The conjugacy of the last layer of a BNN in regression problems was utilized by \textcite{harrison_variational_2024}, who use a Matrix-Normal-Inverse-Wishart distribution to estimate the parameters in the last layer of a Bayesian last layer neural network (BLL, as defined previously). However, when using variational feature learning as we are, this previous work uses MAP estimates of $\bm{\Sigma}$ instead of the conjugate Inverse-Wishart posterior, and does not include a procedure for training using partially-missing response vectors, $\bm{y}$, as we describe in Section \ref{sec:estimation-missing-data}.

With the full conditional distributions in \eqref{eqn:sigma-full-conditional} and \eqref{eqn:weight-full-conditional}, we then  use a conditional last layer variational approximation,
\begin{equation}
    Q(\bm{W}, \bm{b}, \bm{\Sigma}, \bm{\Phi}) \coloneq Q'(\bm{\Phi})p(\bm{W}, \bm{b}, \bm{\Sigma} | \bm{\Phi}, \mathcal{D}), \label{eqn:variational-dist-general}
\end{equation}
with $Q'$ being independent Gaussian distributions. In other words, the posterior distribution of every weight, $w_{ijk}$, in the matrix $\bm{W}_k$ for $k=0,\dots,\ell-1$, and every bias, $b_{jk}$, in the vector $\bm{b}_k$ for $k=0,\dots,\ell-1$ is approximated as a normal distribution $\mathrm{N}(\mu_{w_{ijk}},\sigma_{w_{ijk}})$ or $\mathrm{N}(\mu_{b_{jk}},\sigma_{b_{jk}})$, respectively. The values $\mu_{w_{ijk}},\sigma_{w_{ijk}}, \mu_{b_{jk}},\sigma_{b_{jk}}$ are found by minimizing the KL-divergence of the approximate distribution \eqref{eqn:variational-dist-general} to the true posterior, or equivalently, maximizing the ELBO as is the goal of variational inference. There are no values to be optimized in layer $k=\ell$, as the variational distribution of those weights, biases, and the covariance matrix $\Sigma$ are determined by \eqref{eqn:sigma-full-conditional} and \eqref{eqn:weight-full-conditional} with parameters \eqref{eqn:full-conditional-params-start}-\eqref{eqn:full-conditional-params-end}.

The conjugate form of variational posterior minimizes KL-divergence in the following sense. Consider two distributions, $P(\bm{\theta},\bm{\psi}) = P(\bm{\theta})P(\bm{\psi}|\bm{\theta})$ and $Q(\bm{\theta},\bm{\psi}) = Q(\bm{\theta})Q(\bm{\psi}|\bm{\theta})$, with densities $p$ and $q$, respectively. We consider $P$ fixed and $Q$ as an approximation to $P$, such as in variational inference. The KL divergence of $Q$ from $P$ is:
\begin{align}
    D_{KL}(Q\|P) &= \iint q(\bm{\theta},\bm{\psi})\log\left(\frac{q(\bm{\theta},\bm{\psi})}{p(\bm{\theta},\bm{\psi})}\right) \operatorname{d}\bm{\psi}\operatorname{d}\bm{\theta} \label{eqn:kl-minimization-start}\\
    &= \iint q(\bm{\theta})q(\bm{\psi}|\bm{\theta})\log\left(\frac{q(\bm{\theta})q(\bm{\psi}|\bm{\theta})}{p(\bm{\theta})p(\bm{\psi}|\bm{\theta})}\right) \operatorname{d}\bm{\psi}\operatorname{d}\bm{\theta} \\
    &= \iint q(\bm{\theta})q(\bm{\psi}|\bm{\theta})\log\left(\frac{q(\bm{\theta})}{p(\bm{\theta})}\right) \operatorname{d}\bm{\psi}\operatorname{d}\bm{\theta} \nonumber \\
    &\quad\quad + \iint q(\bm{\theta})q(\bm{\psi}|\bm{\theta})\log\left(\frac{q(\bm{\psi}|\bm{\theta})}{p(\bm{\psi}|\bm{\theta})}\right) \operatorname{d}\bm{\psi}\operatorname{d}\bm{\theta} \\
    &= \underbrace{D_{KL}(Q(\bm{\theta})\|P(\bm{\theta}))}_{(A)} + \underbrace{\int q(\bm{\theta})D_{KL}(Q(\bm{\psi}|\bm{\theta})\|P(\bm{\psi}|\bm{\theta}))\operatorname{d}\bm{\theta}}_{(B)}. \label{eqn:kl-conditional-decomposition}
\end{align}

Term (B) in Equation \eqref{eqn:kl-conditional-decomposition} is the expectation of $D_{KL}(Q(\bm{\psi}|\bm{\theta})\|P(\bm{\psi}|\bm{\theta}))$ with respect to $Q(\bm{\theta})$. Because $D_{KL}(Q(\bm{\psi}|\bm{\theta})\|P(\bm{\psi}|\bm{\theta})) \geq 0$ for all $\bm{\theta}$, term (B) is minimized when $D_{KL}(Q(\bm{\psi}|\bm{\theta})\|P(\bm{\psi}|\bm{\theta})) = 0$ for all $\bm{\theta}$, which is only true when $Q(\bm{\psi}|\bm{\theta}) = P(\bm{\psi}|\bm{\theta})$. Therefore, when performing variational inference on a model with a subset of parameters, $\bm{\psi}$, whose full conditional distribution based on the other parameters, $\bm{\theta}$, and the data, $\mathcal{D}$, is analytically tractable, a better approximation to $P(\bm{\theta},\bm{\psi}|\mathcal{D})$ can always be achieved using an approximate distribution $Q(\bm{\theta})P(\bm{\psi}|\bm{\theta},\mathcal{D})$ instead of an arbitrary $Q(\bm{\theta},\bm{\psi})$, especially a separable $Q(\bm{\theta})Q(\bm{\psi})$.

Our BNN variational posteriors fit this framework with $\bm{\theta} := \bm{\Phi}$, $\bm{\psi} = (\bm{W}, \bm{b}, \bm{\Sigma})$, and all distributions are conditioned on $\mathcal{D}$. Because the distribution \eqref{eqn:last-layer-full-conditional-general} has a known closed form, using \eqref{eqn:variational-dist-general} as the variational distribution provides a better approximation to the true posterior than any other variational distribution using the same $Q'(\bm{\Phi})$. It also simplifies the optimization problem by reducing the complexity of the distribution to be fit to just $Q'$.

\subsection{Conjugate Last Layers with Missing Data} \label{sec:estimation-missing-data}

The approach outlined in the previous section only works when the full conditional distribution, $p(\bm{W}, \bm{b}, \bm{\Sigma} | \bm{Z}, \bm{Y})$, has a closed form. However, if some of the target variables, $\bm{y}_i$, are only partially observed, no such closed-form distribution exists in general. The main case when missing data arises is in the joint model. Due to the concatenation of target values $\bm{y}, \bm{y}^{(1)}, \dots, \bm{y}^{(M)}$ into $\bm{y}'$, any $\bm{x}$ values where only some of the modalities have been observed will have a corresponding $\bm{y}'$ with missing values. For the motivating example of BO, we expect missing data to be a common phenomenon in practice due to not all data sources being observed at the same input values as the algorithm progresses. Missing observations result in partially observed response vectors in the joint model, where any $\bm{x}$ where not all modalities have been observed will have a corresponding joint $\bm{y}$ with missing values.

In an MCMC context, missing data can be sampled along with the parameters using data augmentation \parencite{tanner_calculation_1987}. For monotone missingness and specific prior distributions, \textcite{liu_bayesian_1996} provided a closed-form posterior for regression parameters in multivariate Bayesian linear regression, but they use data augmentation for arbitrary patterns of missingness. Here, we develop an approach for non-monotone missingness and without MCMC or data augmentation, in which we jointly estimate the posterior distribution of the parameters and the posterior predictive distribution of missing data.

Split the response data, $\bm{Y}$, into observed data $\bm{Y}_{obs}$ and missing data $\bm{Y}_{miss}$. If we could draw the missing data from its posterior predictive distribution, $p(\bm{Y}_{miss}|\bm{Y}_{obs},\bm{Z})$, then we could use it along with $\bm{Y}_{obs}$ to produce the desired full conditional distribution:
\begin{equation}
    p(\bm{W}, \bm{b}, \bm{\Sigma} | \bm{Z}, \bm{Y}_{obs}) = \int p(\bm{W}, \bm{b}, \bm{\Sigma} | \bm{Z}, \bm{Y}_{obs}, \bm{Y}_{miss})p(\bm{Y}_{miss}|\bm{Y}_{obs},\bm{Z}) \operatorname{d} \bm{Y}_{miss},
\end{equation}
however, this distribution also does not have a closed form. However, this line of thinking suggests a potentially suitable approximation. Consider splitting the last layer's multivariate regression problem with $k$-dimensional response into $k$ individual Bayesian regression problems with common predictors, $\bm{Z}$. Let $\bm{y}_{j,obs}$ and $\bm{y}_{j,miss}$ be the observed and missing values of the $j^\mathrm{th}$ column of $\bm{Y}$, respectively, and let $\bm{w}_j$ be the $j^\mathrm{th}$ row of $\bm{W}$. Each posterior predictive distribution $p(\bm{y}_{j,miss}|\bm{Z},\bm{y}_{j,obs})$ does have a closed form.

We can then say,
\begin{align}
    p(\bm{W}, \bm{b}, \bm{\Sigma} | \bm{Z}, \bm{Y}_{obs}) &= \int p(\bm{W}, \bm{b}, \bm{\Sigma} | \bm{Z}, \bm{Y}_{obs}, \bm{Y}_{miss})p(\bm{Y}_{miss}|\bm{Y}_{obs},\bm{Z}) \operatorname{d} \bm{Y}_{miss} \\
    &\approx \int p(\bm{W}, \bm{b}, \bm{\Sigma} | \bm{Z}, \bm{Y}_{obs}, \bm{Y}_{miss})\prod_{j=1}^kp(\bm{y}_{j,miss}|\bm{y}_{j,obs},\bm{Z}) \operatorname{d} \bm{Y}_{miss}, \label{eqn:full-conditional-missing-approximate}
\end{align}
and all distributions in \eqref{eqn:full-conditional-missing-approximate} have closed forms. In practice, computing the integral in \eqref{eqn:full-conditional-missing-approximate} is still difficult. 
We instead propose to estimate the joint posterior distribution of the parameters and the missing data,
\begin{align}
    &p(\bm{W}, \bm{b}, \bm{\Sigma}, \bm{\Phi}, \bm{Y}_{miss} | \bm{Y}_{obs}, \bm{X}) \nonumber \\
    &\propto p(\bm{Y}_{miss}|\bm{Y}_{obs},\bm{W}, \bm{b}, \bm{\Sigma}, \bm{\Phi},\bm{X})p(\bm{Y}_{obs}|\bm{W}', \bm{\Sigma}, \bm{\Phi},\bm{X})p(\bm{W}')p(\bm{\Sigma})p(\bm{\Phi}). \label{eqn:posterior-with-missing-data}
\end{align}
\sloppy Because the residual distribution is Gaussian, the missing data's distribution, $p(\bm{Y}_{miss}|\bm{Y}_{obs},\bm{W}, \bm{b}, \bm{\Sigma}, \bm{\Phi},\bm{X})$, is known and we can calculate the unnormalized posterior density in \eqref{eqn:posterior-with-missing-data}. Then for the variational distribution, $Q$, we use the approximate last layer from \eqref{eqn:full-conditional-missing-approximate} to create
\begin{align}
    &Q(\bm{W}, \bm{b}, \bm{\Sigma}, \bm{\Phi}, \bm{Y}_{miss}) \nonumber \\
    &\coloneq Q'(\bm{\Phi})p(\bm{W}, \bm{b}, \bm{\Sigma} | \bm{\Phi},\bm{X}, \bm{Y}_{obs}, \bm{Y}_{miss})\prod_{j=1}^kp(\bm{y}_{j,miss}|\bm{y}_{j,obs},\bm{\Phi},\bm{X}) \label{eqn:variational-q-with-missing-data} \\
    &\approx Q'(\bm{\Phi})p(\bm{W}, \bm{b}, \bm{\Sigma} | \bm{\Phi},\bm{X}, \bm{Y}_{obs}, \bm{Y}_{miss})p(\bm{Y}_{miss} | \bm{Y}_{obs}, \bm{\Phi}, \bm{X}) \\
    &= Q'(\bm{\Phi})p(\bm{W}, \bm{b}, \bm{\Sigma}, \bm{Y}_{miss} | \bm{\Phi},\bm{X}, \bm{Y}_{obs}).
\end{align}
Each component of \eqref{eqn:variational-q-with-missing-data} has a known closed form that can be sampled, allowing its use in SVI.

\section{Simulations and Empirical Evaluation} \label{sec:simulation}

%\julie{it would help somewhere to express what are the input and output in the following, especially when you transition to time series }

To demonstrate the effectiveness of these models, we apply them to simulated and real-world datasets and measure both their in-sample and out-of-sample predictive ability with prediction bias and standardized error. The following subsections describe the data, experiments, and results.

\subsection{Simulated Data} \label{sec:simulation-data}

We test our developments on eight total configurations of four distinct datasets, each comprising of a main and auxiliary modalities. A full description of each dataset is available in Appendix \ref{apx:simulated-datasets}.
\begin{itemize}
    \item Branin: the Branin function \parencite{surjanovic_virtual_2013}, augmented with ``low fidelity'' versions of varying correlation with the main function \parencite{toal_considerations_2015}. The input is two-dimensional and the output of each function is a scalar. The unmodified function is the main modality.
    \item Paciorek: the Paciorek function augmented with ``low fidelity'' versions of varying correlation with the main function \parencite{toal_considerations_2015, mainini_analytical_2022}. The input is four-dimensional and the output of each function is a scalar. The unmodified function is the main modality.
    \item Paciorek (high): the above multi-fidelity Paciorek functions, with only low fidelity functions that have high correlation with the main function.
    \item Paciorek (low): the above multi-fidelity Paciorek functions, with only low fidelity functions that have low correlation with the main function.
    \item Wind: The wind dataset is built from wind time series extracted from the ERA-5 model dataset \parencite{hersbach_era5_2020} and measurements from Argonne National Laboratory tower measurements accessible at \url{https://www.anl.gov/evs/atmos}  comprising hourly wind speed and direction. For these data the input is 1-dimensional (time) and the output of each modality is scalar. Measured wind speed at 10 meters is the main modality, and measurements at other altitudes, and modeled wind speed and direction are the auxiliary modalities. This illustrates a missing data imputation procedure where missing observations are infilled from regularly available model data and other observations. In practice, measurements are frequently missing, linear interpolation between missing data without additional sources of information is commonly used in these situations. However, this is poorly informative and fully deterministic. 
    \item Wind (daily): Using the same wind data, we consider the input as discrete days instead of hours. Thus, each observation of a modality is a 24-dimensional vector which we reduce in dimension through PCA before modeling.
    \item Time Series: a synthetic dataset consisting of noisy time series data and several summaries of the entire series computed from the time series without noise. The input is three variables that control the overall shape of the time series. We use a PCA-reduced version of the full time series as the main modality.
    \item Time Series (1d): This dataset is the same synthetic dataset as above, but considering one scalar summary of the time series as the main quantity of interest and the PCA-reduced time series, and other summaries as auxiliary modalities.
\end{itemize}
%Two datasets are standard multi-fidelity Bayesian optimization test functions \parencite[denoted Paciorek and Branin;][]{mainini_analytical_2022, surjanovic_virtual_2013}, with low fidelity versions of the functions treated as multi-modal data. One dataset is a function built from wind time series extracted from the ERA-5 model dataset \parencite[denoted Wind;][]{hersbach_era5_2020} and from measurements from \julie{i will add a ref here} comprising hourly wind speed and direction. Lastly, we use two configurations of a synthetic dataset consisting of noisy time series data and several summaries of the entire series computed from the time series without noise (denoted Time Series in the following). We construct datasets by choosing a main modality for each and sampling that function at a relatively small number of points (36 to 256) along with a larger number of points of the alternate modalities. For the synthetic time series dataset, we alternately select the \todo{(PCA-decomposed time series itself as the main modality or one of the summary functions as the main modality (Time Series (1d))}.
These data sets were chosen to have a variety of input dimensions, output dimensions, and both the number and the informative power of auxiliary modalities. For all datasets, the main modality was sampled at a sparse grid of input values and auxiliary modalities were sampled at a superset of input values including locations both inside and outside of the convex hull of the main modality training data. This choice was made to better emulate real-world settings where auxiliary data is easier to acquire than direct observations of the quantity of interest (main modality). The set of input values for model evaluation was chosen to be the set of all input values with auxiliary modality data. We standardize the training and validation data by subtracting the mean of the training data and dividing by the standard deviation of the training data.

%\julie{do you wnat to have a separate paragarph about the dimension reduction since it is also a modeling strategy and it is a very nice feature of your model that i think deserves much more highlights, say that you output the coefficients and reconstruct evyerythign and bla bla  }
High-dimensional data presents a challenge for our proposed models, due to the presence of the Inverse-Wishart covariance matrix in \eqref{eqn:bnn-priors-Sigma}. For an output with dimension $k$, this parameter will have dimension $k\times k$, and it is difficult to calculate the conjugate Wishart parameters and sample this matrix without numerical error for large $k$. Therefore, when modeling high-dimensional data, we first reduce the dimension through PCA. We choose PCA due to its simplicity, speed and effectiveness, in terms of limited need for tuning, its ease of decomposition and reconstruction of the data, and the resulting reduction in data size in our test datasets. Other dimension reduction methods could also be used. For a high-dimensional modality $\bm{y}^{(m)}$, we perform a PCA analysis on all observations of $\bm{y}^{(m)}$ in the training data. Then, retaining the first $c$ largest components to explain at least 95\% of the variance in the data, we replace each observation $\bm{y}^{(m)}$ with a $c$-dimensional vector $\bm{y}^{(m)}_{PCA}$ of the coefficients of these $c$ components. The vectors $\bm{y}^{(m)}_{PCA}$ are used in model training. For model evaluation, we project the validation points of that modality onto the same space of principal components, keeping the same $c$ coefficients, and compare to the model's predictions in the reduced space.  Dimension reduction was necessary for the target variables in the Time Series, Time Series (1d), and Wind (daily) datasets. The Time Series dataset's high-dimensional time series modality was reduced from length 200 to length 7. For the ``Wind (Daily)'' dataset, while each target variable only has dimension 24, there are 17 total variables of this size used, for a total size of 408 dimensions across all target variables. We reduce the dimension of each target variable using PCA to between 3 and 10, for a total of 94 dimensions across all 17 modalities.

\subsection{Numerical Experiment Framework} \label{sec:simulation-framework}

For each dataset, we fit a joint model (Joint; Section \ref{sec:model-joint}) and a layered model (Layered; Section \ref{sec:model-layered}) to the full multi-modal data. We also fit a single BNN of the form in \eqref{eqn:neural-net-start}-\eqref{eqn:neural-net-end} on just the main modality data (Single). Comparing each model to the Single BNN allows for benchmarking our multi-modal surrogates against the most similar uni-modal BNN surrogate available. Each model was fit 20 times using different initial seeds to account for the stochastic nature of the estimation procedure. When constructing the training data for the joint model, we assumed the target variable defined in \eqref{eqn:augmented-response-variable}, $\bm{y}'$, had missing values wherever the auxiliary modalities were observed but the main modality was not. These missing values were handled as described in \ref{sec:estimation-missing-data}. Each neural network comprising the models had two hidden layers of 256 nodes each. We qualitatively saw negligible change in performance for deeper or wider networks, thus, selected 2 layers of 256 nodes as the architecture for these experiments. Each model was randomly initialized and trained until there was no statistically significant slope in the loss function over each epoch. After training, we drew 500 samples of the parameters from the trained variational posterior to use to evaluate the model's predictive performance.

Several metrics are computed on the predictions from these models to quantitatively assess their performance. For each dataset and fitted model, we calculated the prediction bias and standardized error for in-sample datapoints (Sample), out-of-sample test datapoints inside the convex hull of the main modality training data (In Hull), and out-of-sample test datapoints outside the convex hull (Out of Hull). The performance at test datapoints inside the convex hull of the training data measures the models' ability to interpolate. This performance is relevant in downstream tasks such as BO, where surrogate models are often trained on an initial experimental design that covers the area of the input space where the optimal point is expected to be found. The performance at test datapoints outside of the convex hull of the training data, measures the models' ability to extrapolate. This performance is relevant when auxiliary data is available for a wider portion of the domain than main modality data, which happens when the main quantity of interest is more expensive to collect or simulate.
 
The bias for a point $(\bm{x}^\ast,\bm{y}^\ast)$ was calculated using the formula,
\begin{equation}
    \mathrm{Bias} = \|\bm{y}^\ast - \hat{\bm{\mu}}^\ast\|_2,
\end{equation}
where $\hat{\bm{\mu}}^\ast = \frac{1}{n}\sum_{i=1}^n\bm{\mu}_i^\ast$ is the posterior mean estimate over the $n=500$ posterior samples of $\bm{\mu}$ at the point $\bm{x}^\ast$. Low bias indicates that model predictions are accurate, on average, to the true unobserved function values.

The standardized error was calculated using the formula,
\begin{equation}
    \left(\frac{1}{p}(\bm{y}^\ast - \hat{\bm{\mu}}^\ast)^\top \bm{V}^{\ast-1}(\bm{y}^\ast - \hat{\bm{\mu}}^\ast)\right)^{-1/2}, \label{eqn:standardized-error}
\end{equation}
where $\bm{V}$ is the $k\times k$ estimated posterior covariance from the 500 samples of $\bm{\mu}$ at $\bm{x}^\ast$, and $\bm{\mu}$ and $\bm{y}^\ast$ have dimension $k$. Equation \eqref{eqn:standardized-error} standardizes the error in the following sense: if $\bm{y}\sim\mathrm{N}_p(\bm{\mu},\bm{\Sigma})$, then $(\bm{y}-\bm{\mu})^\top\bm{\Sigma}^{-1}(\bm{y}-\bm{\mu}) \sim \chi^2_k$ and $\mathrm{E}\left[(\bm{y}-\bm{\mu})^\top\bm{\Sigma}^{-1}(\bm{y}-\bm{\mu})\right] = k$. Therefore, if the model is properly calibrated and the residuals follow a normal distribution, the square of \eqref{eqn:standardized-error} approximately follows a scaled $\chi^2$-distribution, and \eqref{eqn:standardized-error} approximately follows a scaled $\chi$-distribution. We add the square root in \eqref{eqn:standardized-error} to reign in the potentially long tails of the $\chi^2$-distribution and allow for a better comparison between models. An expectation of the standardized error over all test points that is very far away from 1 indicates either bias in the posterior mean, poorly-calibrated posterior variance, or both.

\subsection{Numerical Experiment Results} \label{sec:simulation-results}

\begin{figure}
    \centering
    \includegraphics[width=\linewidth]{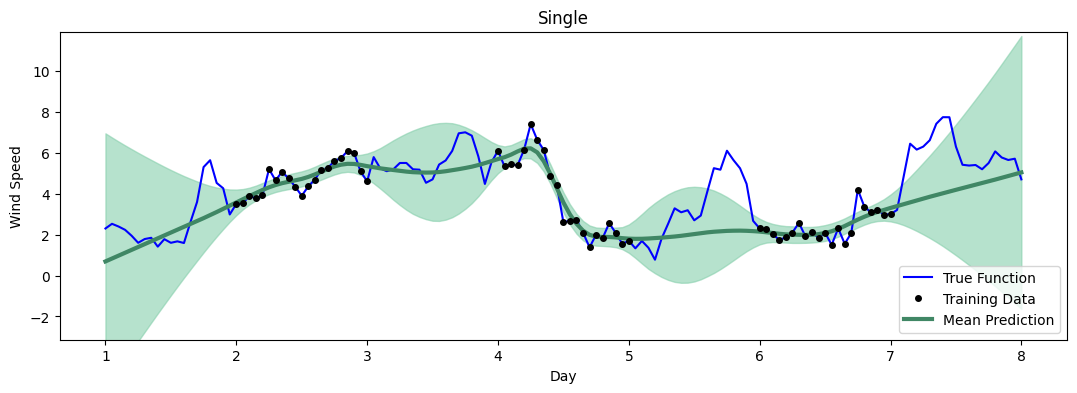} \\
    \includegraphics[width=\linewidth]{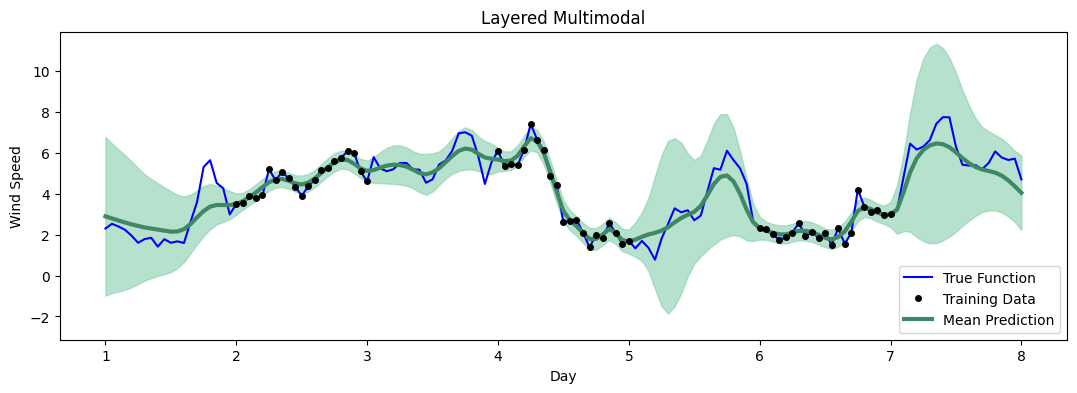}
    \caption{Example of uni-modal (top) and Layered multi-modal (bottom) models fit to the same wind dataset. Notice the visible reduction in average prediction error (visualized as the difference between the blue and green lines) in the layered multi-modal model where there are no observations of the main quantity of interest, relative to the uni-modal model. Because auxiliary data was used in training in the gaps between main modality training data (see Figure \ref{fig:wind-aux-training-data}), the mean prediction is able to more closely follow the true function in those areas, the posterior prediction intervals (visualized as the light green bands) more often captures the true function.}
    \label{fig:results-example}
\end{figure}

\begin{figure}
    \centering
    \includegraphics[width=\linewidth]{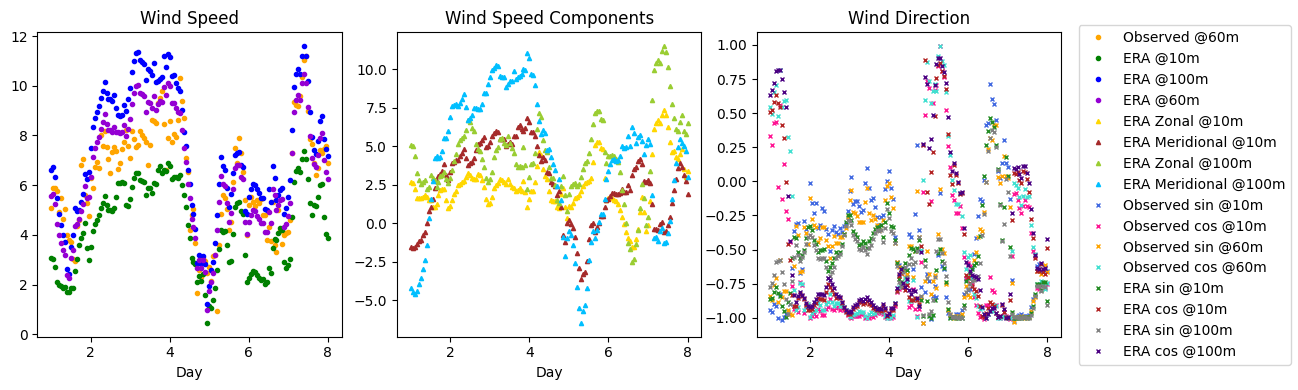}
    \caption{Auxiliary data that was used to train the layered multi-modal model on the Wind dataset. The auxiliary data is measured both where the main modality is measured and also where the main modality is not. Therefore the auxiliary network is able to assist the main network in making predictions where there is no main modality training data.}
    \label{fig:wind-aux-training-data}
\end{figure}

\begin{figure}
    \centering
    \includegraphics[width=\linewidth]{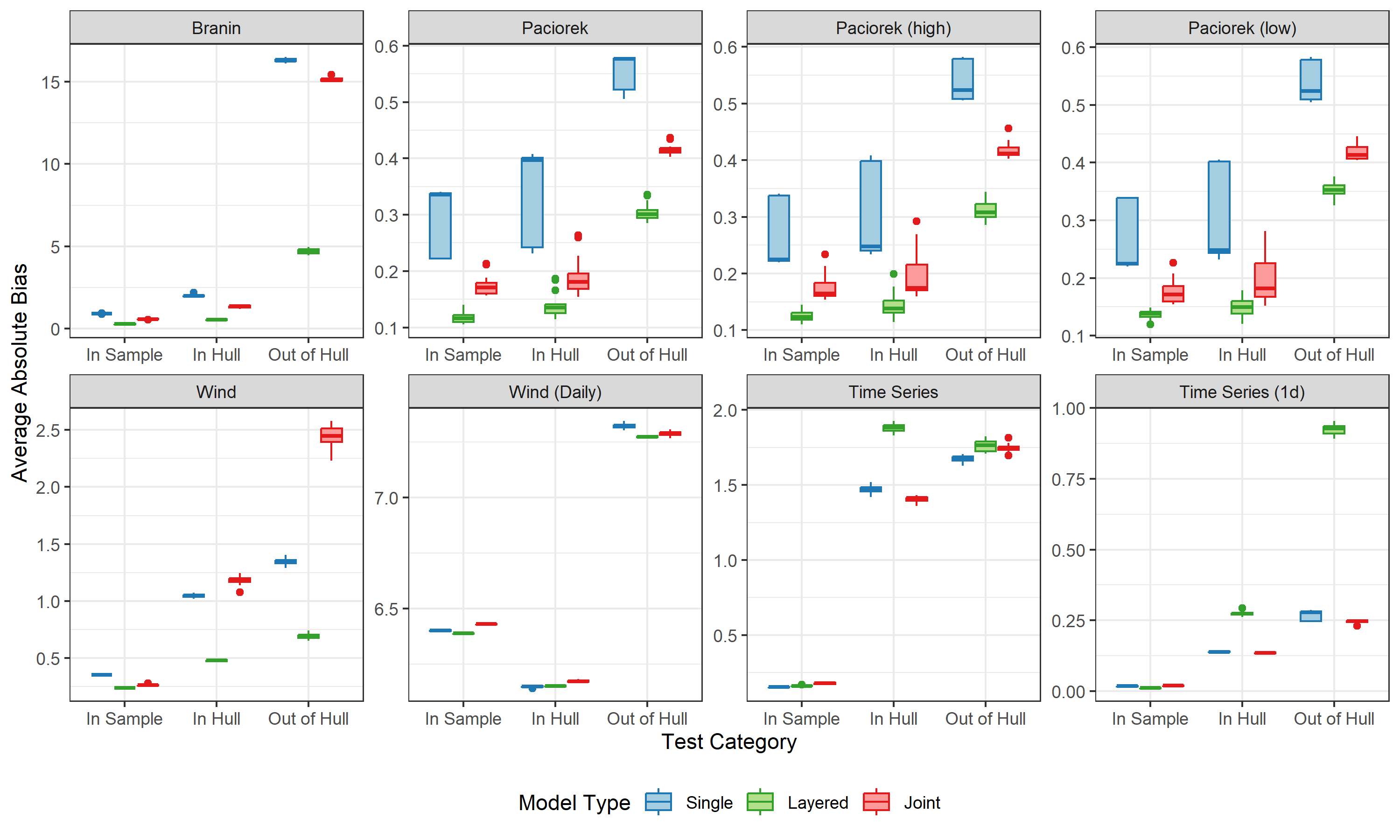}
    \caption{Average bias for multi-modal models on in-sample and out-of-sample predictions compared to uni-modal models. Low bias indicates that model predictions are accurate, on average, to the true unobserved function values.}
    \label{fig:results-bias}
\end{figure}

\begin{figure}
    \centering
    \includegraphics[width=\linewidth]{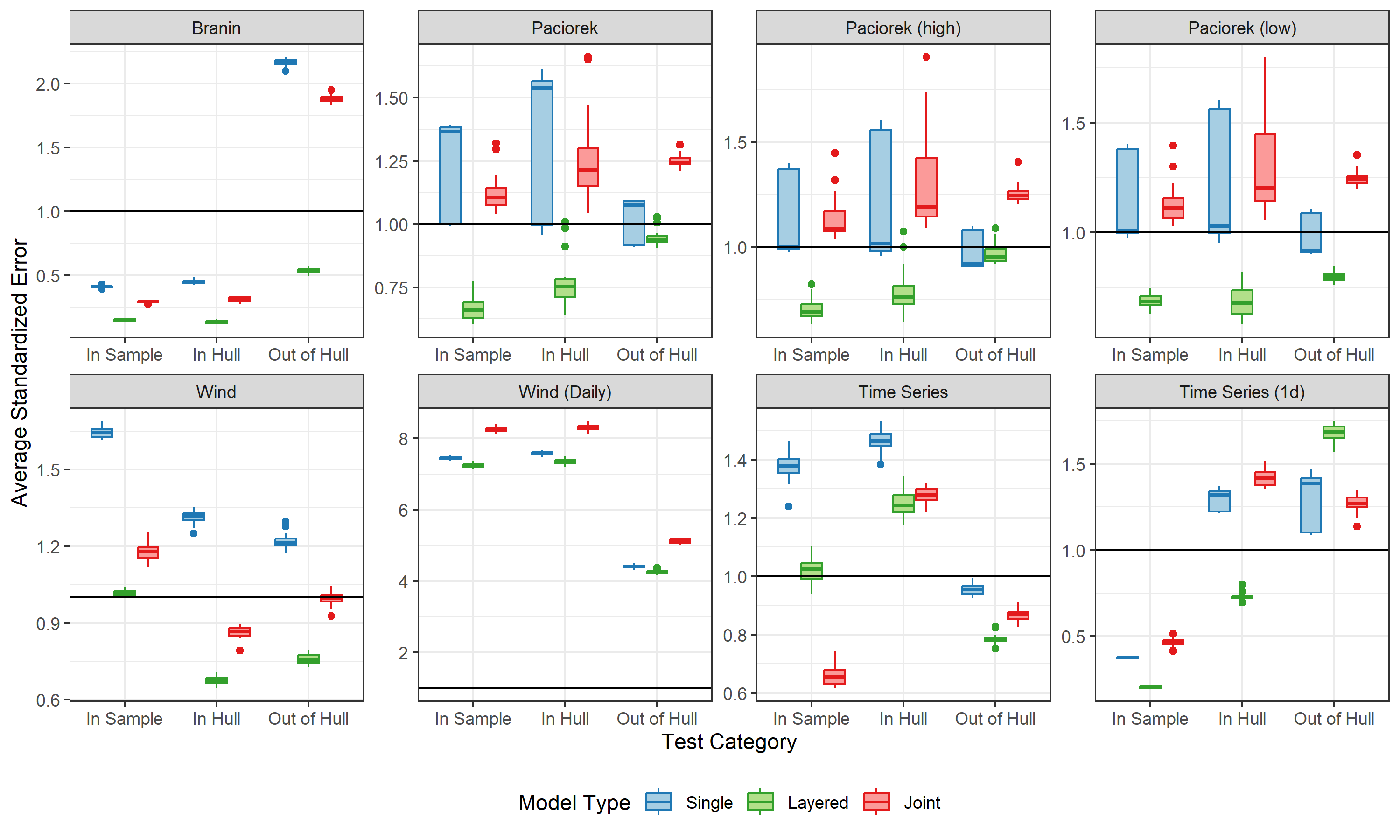}
    \caption{Average standardized error for multi-modal models on in-sample and out-of-sample predictions compared to unimodal models. An expectation of the standardized error over all test points that is very far away from 1 indicates either bias in the posterior mean, poorly-calibrated posterior variance, or both.}
    \label{fig:results-zscore}
\end{figure}

The multi-modal surrogate models show promising improvement on some datasets over the uni-modal model. For both Branin and Paciorek datasets, the layered model greatly reduces the bias of the predictions compared to the uni-modal model, and the joint model slightly reduces bias (Figure \ref{fig:results-bias}). The difference is particularly noticeable for test points outside of the convex hull of the main modality training data. The multi-modal models also perform at least as well as the uni-modal model in terms of standardized error, which we expect to be closer to one if model uncertainty is well-calibrated (Figure \ref{fig:results-zscore}). Interestingly, both multi-modal models perform very similarly in all Paciorek datasets, regardless of the correlation of the auxiliary modalities. The layered model shows similar results on the wind dataset, while the joint multi-modal model has higher prediction bias (but better-calibrated errors) than the uni-modal model for out-of-sample test points. Figure \ref{fig:results-example} shows one trained instance of the layered multi-modal model improving out-of-sample predictions on the wind dataset due to the presence of multi-modal data in the gaps between the main modality training points (Figure \ref{fig:wind-aux-training-data}). In this fitted example, the auxiliary training data affects both the mean prediction and the prediction uncertainty. The mean prediction follows the true mean more closely where there is no main modality training data. In areas where the auxiliary training data is less variable ($t=3$ to $t=4$), the prediction uncertainty is decreased, while in areas where the auxiliary training data is more variable ($t=5$ to $t=6$), the uncertainty is increased. The layered model improves prediction bias and standardized error in the Wind (Daily) dataset, although both measures of error are overall higher than in other datasets. On both Time Series datasets, the layered multi-modal model performs worse in terms of average bias than both the uni-modal and joint multi-modal models, which perform comparably to each other. On the Time Series (1d) dataset, the layered model appears to have standardized errors closer to one for the In Sample and In Hull test points, but farther from one for the Out of Hull test points. \\

To investigate differing informative power of the auxiliary modalities in each dataset that lead to differing performance of the multi-modal models, we calculate the canonical correlation between the quantity of interest and the auxiliary modalities. Canonical correlation is the maximal linear correlation between any possible linear combinations of two random vectors,
$\mathrm{ccorr}(\bm{X},\bm{Y}) = \underset{\bm{a},\bm{b}}\max\ \mathrm{corr}(\bm{a}^\top\bm{X},\bm{b}^\top\bm{Y})$. We treat values of the functions at different input locations as independent observations of a random variable, and consider all points in the validation dataset of our data. This approach was used by \textcite{toal_considerations_2015} in the context of co-Kriging and multiple fidelity surrogate models to measure the linear correlation and RMSE between individual scalar low-fidelity sources and high-fidelity sources. Canonical correlation is more general than linear correlation and RMSE, allowing the association between two random vectors to be measured in a single scalar. Higher canonical correlation between the main modality and auxiliary modalities should allow for the auxiliary modalities to lend more predictive performance increase to the multi-modal models over their uni-modal counterparts. The results of these calculations are gathered in Table \ref{tab:data-correlations}.
\begin{table}
    \caption{Each dataset's estimated canonical correlation between the main quantity of interest (modality) and auxiliary modalities. Canonical correlation exists on a scale from zero to one, and higher canonical correlation values indicate stronger multi-linear relationships between the auxiliary modalities and the main modality.}
\centering
\begin{tabular}{l|r}
\hline
Dataset & Canonical Correlation \\
\hline
Branin & 1.0000\\
\hline
Paciorek & 1.0000\\
\hline
Paciorek (high) & 1.0000\\
\hline
Paciorek (low) & 1.0000\\
\hline
Wind & 0.9828\\
\hline
Wind (Daily) & 0.9701\\
\hline
Time Series & 0.5772\\
\hline
Time Series (1d) & 0.5488\\
\hline
\end{tabular}
\label{tab:data-correlations}
\end{table}
We notice a stark difference in the canonical correlation between the Time Series datasets and other datasets. In particular, for the multi-fidelity functions Branin and Paciorek, we see canonical correlations of exactly one. This is due to the construction of the auxiliary modalities as the main modality plus an offsetting function (see Appendix \ref{apx:simulated-datasets}), and conveys an important property of these data. Together, the auxiliary modalities provide more information than each individually. The consistency of the high canonical correlation with all Paciorek datasets  can also explain why we do not see the decrease in performance of the multi-modal models with lower correlation auxiliary modalities that we may expect to see based on pairwise correlations.

We designed our multi-modal models to account for non-linear relationships, but the strength of linear relationships appears to be important in their ability to learn from auxiliary modalities, and linear relationships are present in the multi-modal models. In the joint multi-modal model, the modalities are related in the last layer via the covariance matrix $\bm{\Sigma}$, and in the layered model, the first layer of the final BNN performs a linear combination of the predicted values of the auxiliary modalities. Canonical correlation analysis between the main and auxiliary modalities can be an informative first step in determining whether these multi-modal models may make better predictions than a uni-modal alternative.

We also explored these datasets using mutual information as a measure of non-linear association between modalities, however mutual information reveals not suitable for this use. Because mutual information measures the amount of ``information'' about one variable that is obtained when observing another, the mutual information between two deterministic continuous values such as the auxiliary and main modalities in our data is theoretically infinite, so mutual information would not allow any differentiation between datasets. 

\section{Application Datasets} \label{sec:application}

We apply these models to two real-world datasets: a dataset of building heating and cooling energy demand \parencite[ENB;][]{tsanas_accurate_2012} and a dataset of heavy metal concentrations in the Jura region of Switzerland \parencite[Jura;][]{goovaerts_geostatistics_1997}.

\subsection{ENB Dataset}

The ENB dataset contains heating and cooling demand for 768 generated buildings as simulated by Autodesk Ecotect Analysis. Ecotect is a building performance analysis tool widely used by architects, engineers, and sustainability consultants. \footnote{Autodesk Ecotect Analysis has been discontinued as of March 20, 2015. (\url{https://www.autodesk.com/support/technical/article/caas/sfdcarticles/sfdcarticles/Ecotect-Analysis-Discontinuation-FAQ.html})}.  The buildings were generated with the same overall volume, location, interior characteristics and building materials, but with differing layout configurations, glazing areas and distributions, and cardinal orientations. The input variables recorded were relative compactness, surface area, wall area, roof area, overall height, orientation, glazing area, and glazing area distribution. These building configurations result in 586 unique heating loads and 636 unique cooling loads.

\textcite{tsanas_accurate_2012} use these data to demonstrate the applicability of random forest  regression models over linear regression models for predicting building energy use, finding that random forests outperform linear regression due to the complex input-output relationships in the data. While in the original analysis, both variables are always measured, and there is no difference in computational difficulty between the heating and cooling simulation, one can suppose a different setting where this is not the case. For example, heating demand may be directly measured on buildings in the winter, before the study is expanded and cooling demand is measured on those and more buildings in the summer. In this setting, our multi-modal surrogate models may be trained on this dataset to predict heating demand both on new buildings and on buildings for which only cooling demand was measured. By comparison, waiting until winter to measure heating demand for the new buildings is undesirable. In our analysis, we treat heating demand as the main modality and cooling demand as an auxiliary modality. Because both heating and cooling demand are strictly positive in these data, we log-transform each variable before fitting models. Results are shown for the $\log(\bm{y})$ rather than $\bm{y}$.

We fit models on several smaller subsets of the ENB data. We randomly select 8, 16, 32, 64, 128, or 256 points as main modality training data ($n_{main}$), with 1, 2, or 4 times as much auxiliary modality data ($r_{aux}$). Auxiliary data are as with other datasets, selected at a superset of input values as the main modality. (Therefore datasets with a ratio of 1:1 auxiliary to main modality data will have all training data for the same input values.) For every combination of these values, we generate 10 datasets by selecting different points from the overall data. For each dataset, we fit a Single, Joint, and Layered model (described in Section \ref{sec:simulation-framework}), and record the mean prediction, prediction error, prediction variance, and standardized prediction error for the main modality at each point.

\begin{figure}
    \centering
    \begin{subfigure}{0.49\textwidth}
        \includegraphics[width=\linewidth]{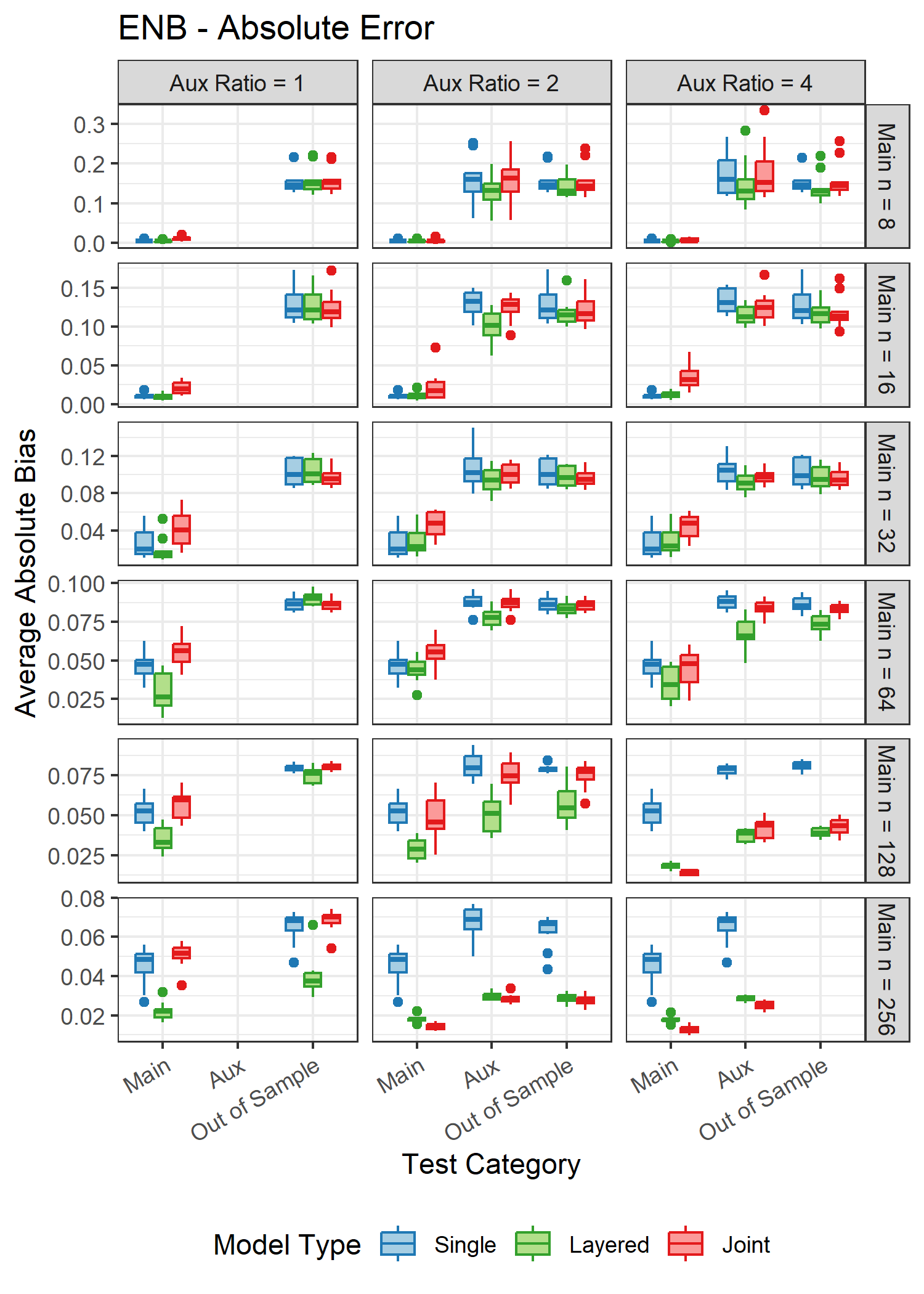}
    \end{subfigure}
    \begin{subfigure}{0.49\textwidth}
        \includegraphics[width=\linewidth]{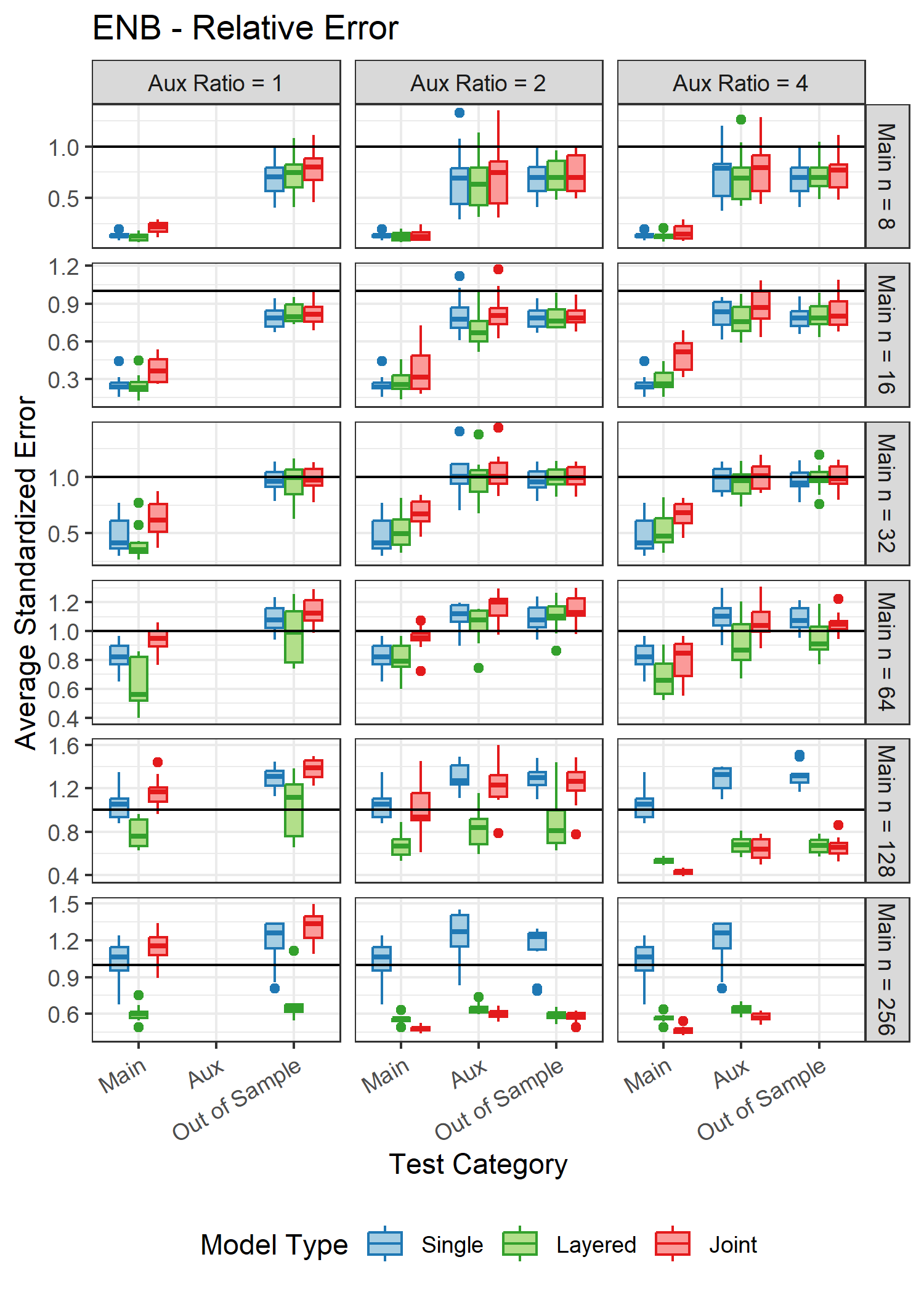}
    \end{subfigure}
    \caption{Comparison of average absolute prediction error (left) and relative prediction error (right) for the ENB dataset, by dataset parameters and test category. The ``Main'' category includes all points where there were main modality (heating load) training data, the ``Aux'' category includes all points where there were not main modality training data but there were auxiliary modality data, and the ``Out of Sample'' category includes all other points. There is no ``Aux'' category where Aux Ratio equals 1, because there were no input values for which only auxiliary modalities were observed in the training data. Additionally, the ``Out of Sample'' category is missing where the training data, either main or auxiliary modality, encompassed the entire dataset (e.g., in the bottom-right facet, $256 \times 4 > 768$, so no points were fully left out of the training data).}
    \label{fig:enb-summary}
\end{figure}

The results of these simulations are summarized in Figure \ref{fig:enb-summary}. We evaluate prediction accuracy on three subcategories of the dataset: Main, Aux, and Out of Sample. The Main category contains in-sample points where the main modality response was observed in the training data. The Aux category contains in-sample points where the main modality response was not observed in the training data, but auxiliary modalities were. Finally, the Out of Sample category contains all other points where no training data were present. The absolute error and standardized error are averaged over every point in each category, and the figure shows the distribution of this category average over the 10 unique simulations for each combination of $n_{main}$, $r_{aux}$, and model type. We are interested primarily in the multi-modal models' ability to make improved predictions at points outside of the main modality training data, either with auxiliary training data present (Aux) or without (Out of Sample). A lower absolute prediction error indicates more accurate predictions, on average, and a standardized prediction error closer to 1 indicates properly-calibrated prediction uncertainty. We see improvement in absolute error for multi-modal models over the uni-modal model for larger values of $r_{aux}$. This trend is expected, as the more auxiliary data are present, the more additional information is available for the multi-modal models to inform their predictions. We also see an improvement in the multi-modal models for larger values of $n_{main}$. This trend is also expected, as when $n_{main}$ increases there is a larger amount of paired main and auxiliary training data the model can use to learn relationships between modalities. Overall, both the Joint and Layered models see improved prediction error. However, the Layered model has the largest and most consistent improvements, showing smaller error than both the Single and Joint models for some medium values of $n_{main}$ and $r_{aux}$. From the standardized error, we see overall well-calibrated uncertainty in the Aux and Out of Sample categories (standardized error close to 1). There is a noticeable reduction in standardized error in the multi-modal models for combinations of both large $n_{main}$ and large $r_{aux}$, however, it remains relatively close to 1.

\subsection{Jura Dataset}

The Jura dataset contains measurements of seven heavy metal concentrations (chromium, nickel, lead, zinc, cadmium, cobalt, and copper) at 359 locations in the Swiss Jura region. In addition to the multiple target variables, each observation contains $X$ and $Y$ geographic coordinates, a categorical variable for one of five rock types, and a categorical variable for one of four land uses. The categorical variables are encoded as five and four indicator variables, respectively. Most observations lie in a geographic grid, with unstructured clusters of additional observations around several points.

This dataset has been analyzed in a multi-target regression context by \textcite{spyromitros-xioufis_multi-target_2016}, who identify that the correlations between measurements can be used to predict concentrations of metals that are more expensive to test for using measurements of metals which are less expensive to test for. For our tests, we use the zinc measurement as the scarce main modality and the other metals as the auxiliary modalities. Because each target variable is strictly positive and right-skewed, we log-transform each variable before fitting models. Results are shown for the $\log(\bm{y})$ rather than $\bm{y}$.

We fit models on several smaller subsets of the JURA data, following the same design as with the ENB data. We randomly select 8, 16, 32, 64, 128, or 256 points as main modality training data ($n_{main}$), with 1, 2, or 4 times as much auxiliary modality data ($r_{aux}$). Certain values of rock type and land use are present in very few points in the dataset. In order to avoid imbalanced data problems, and to simplify the demonstration, we exclude the land use and rock type categorical inputs. The imbalanced data issue could also be resolved in other ways such as oversampling and is beyond the scope of this work. For every combination of these values, we generate 10 datasets by selecting different points from the overall data. For each dataset, we fit a Single, Joint, and Layered model (described in Section \ref{sec:simulation-framework}), and record the mean prediction, prediction error, prediction variance, and standardized prediction error for the main modality at each point.

\begin{figure}
    \centering
    \begin{subfigure}{0.49\textwidth}
        \includegraphics[width=\linewidth]{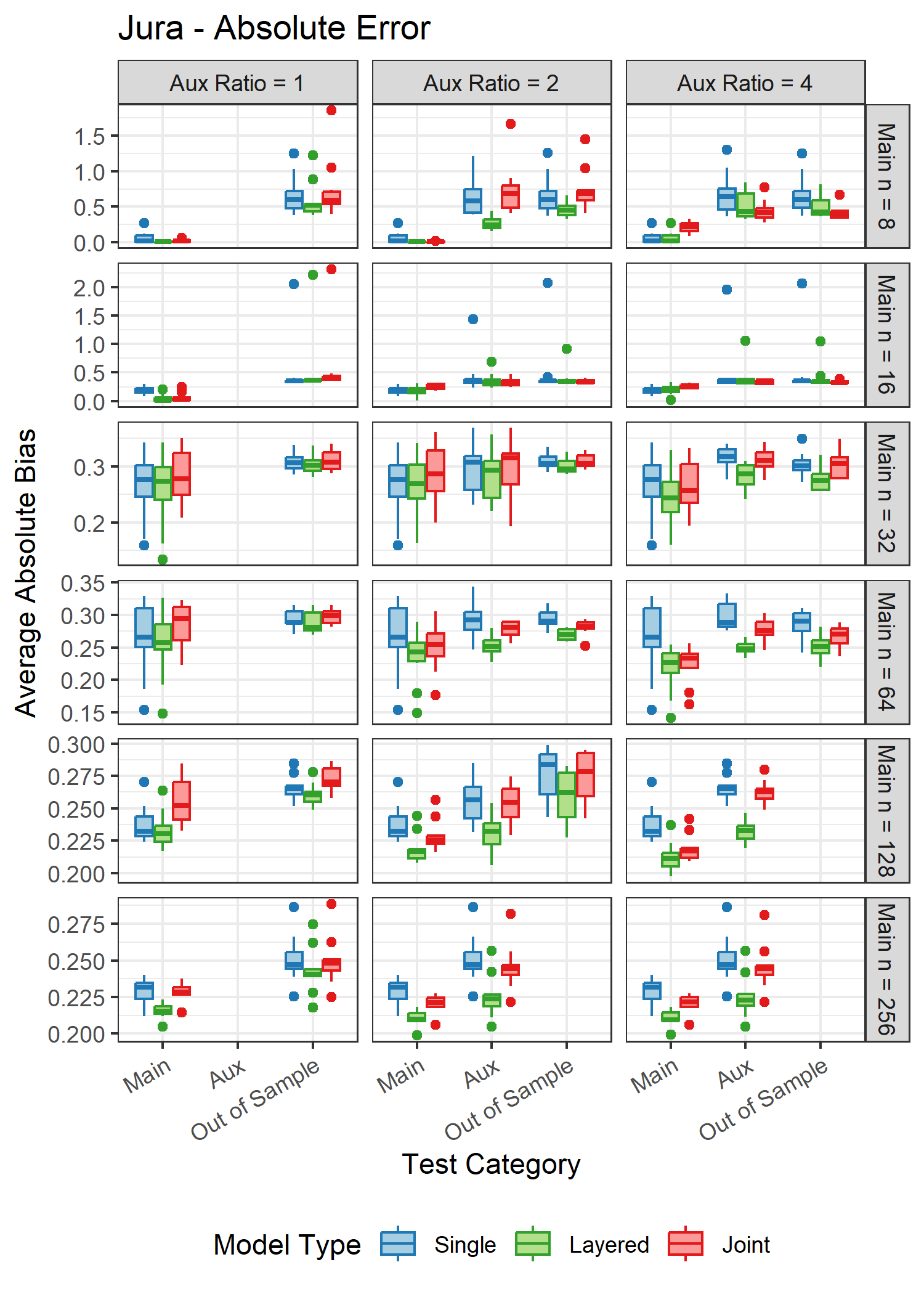}
    \end{subfigure}
    \begin{subfigure}{0.49\textwidth}
        \includegraphics[width=\linewidth]{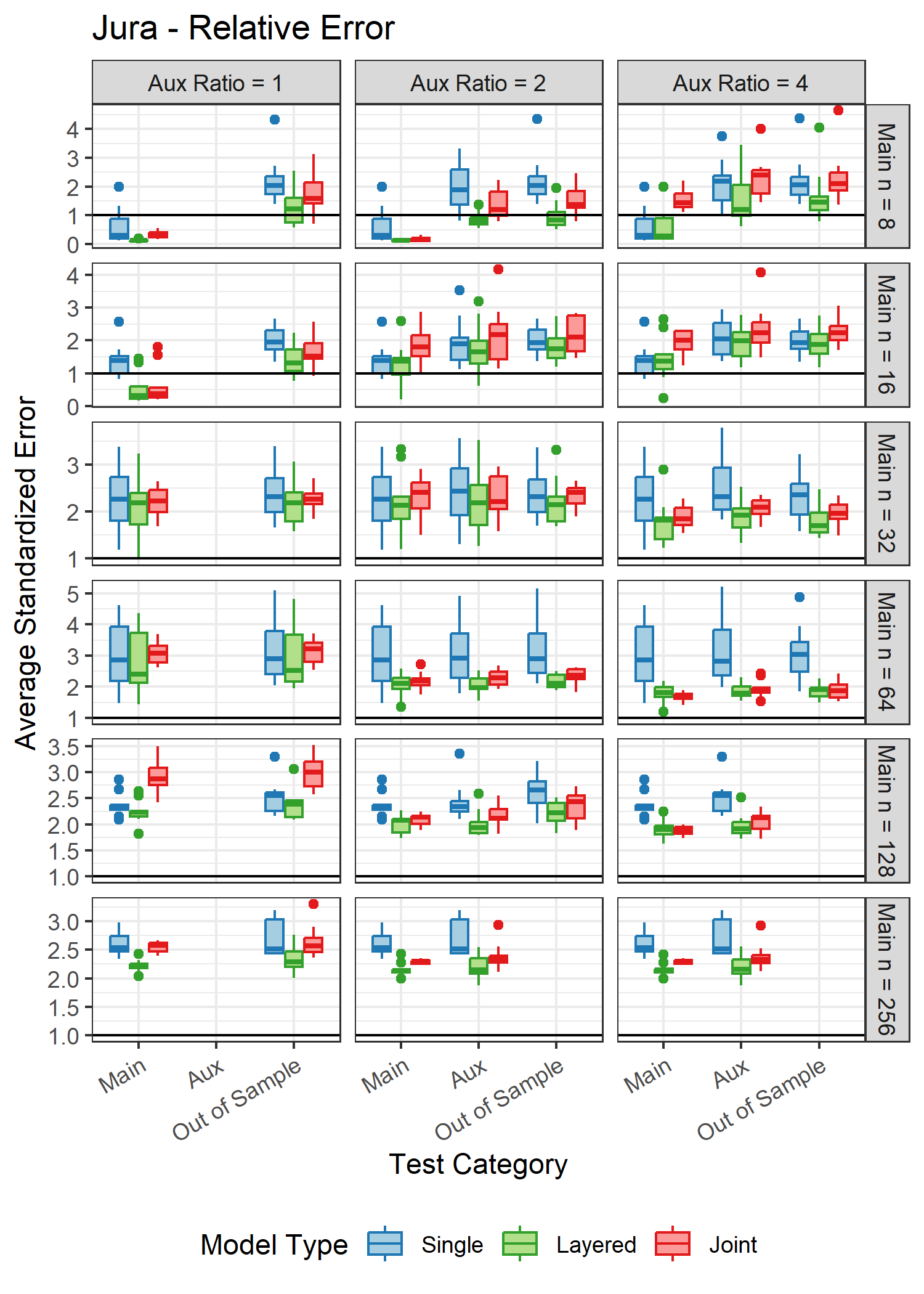}
    \end{subfigure}
    \caption{Comparison of average absolute prediction error (left) and relative prediction error (right) for the Jura dataset, by dataset parameters and test category. The ``Main'' category includes all points where there were main modality (Zinc) training data, the ``Aux'' category includes all points where there were not main modality training data but there were auxiliary modality data, and the ``Out of Sample'' category includes all other points. There is no ``Aux'' category where Aux Ratio equals 1, because there were no input values for which only auxiliary modalities were observed in the training data. Additionally, the ``Out of Sample'' category is missing where the training data, either main or auxiliary modality, encompassed the entire dataset (e.g., in the bottom-right facet, $256 \times 4 > 359$, so no points were fully left out of the training data).}
    \label{fig:jura-summary}
\end{figure}

Figure \ref{fig:jura-summary} shows an overall summary view of model performance on the Jura data, just as Figure \ref{fig:enb-summary} shows on the ENB data. The Main, Aux and Out of Sample categories, and category-wide averages of errors are created in the same manner as with the ENB data. We see some of the same overall trends in average bias as with the ENB data, with multi-modal models showing improvement over uni-modal models for large $n_{main}$ and $r_{aux}$. However, where for the ENB data we saw more comparable improvement between the Joint and Layered models, for the Jura data we see the Layered model outperforming the Joint model in most circumstances. Additionally, the calibration of the uncertainty quantification is generally poorer for the Jura dataset, as shown by the standardized prediction error (right panel) being larger than 1, often between 2 and 3. This indicates that the uncertainty estimate for predictions is generally smaller than it should be if it were properly calibrated. This discrepancy could be due to the fact that the Jura data are measurements, while the ENB data are from simulations. Even though the standardized error is high, it is still closer to 1 for the multi-modal models than for the uni-modal model, especially for larger values of $r_{aux}$.

\begin{figure}
    \centering
    \begin{subfigure}{0.49\textwidth}
        \includegraphics[width=\linewidth]{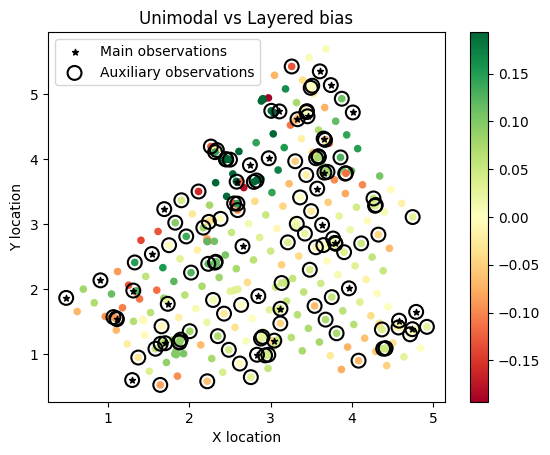}
    \end{subfigure}
    \begin{subfigure}{0.49\textwidth}
        \includegraphics[width=\linewidth]{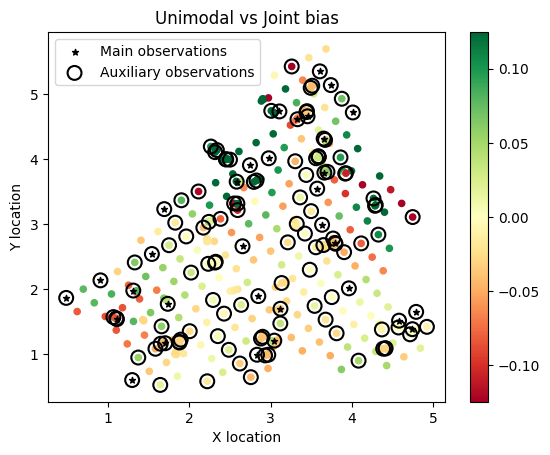}
    \end{subfigure}
    \caption{A comparison of prediction bias (average absolute error) for each multi-modal model vs the uni-modal model on one instance of a trained model with $n_{main} = 32$ and $r_{aux} = 4$. Positive values (green) mean the multi-modal model reduced average error versus the uni-modal model, while negative values (red) mean the multi-modal model increased average error. The locations of main modality training data are shown as stars, and the locations of auxiliary training data are shown as circles. Points co-located with auxiliary data are predominantly green in the Layered model and green or yellow in the Joint model, indicating that predictions were improved relative to the uni-modal model.}
    \label{fig:jura-example-fit-bias}
\end{figure}

We can take advantage of the Jura dataset's two-dimensional spatial input values to visualize multi-modal model performance in one trained instance. Figure \ref{fig:jura-example-fit-bias} shows a comparison of main modality prediction bias for one instance of the multi-modal models compared to the uni-modal model. Qualitatively, we see more green points inside the circles, i.e., where auxiliary modality responses were included in training data but the main modality response was not. We also see in some areas between groups of auxiliary observations, the points are also more often green, indicating that the auxiliary training data improves interpolation to out-of-sample test points.

\section{Conclusion} \label{sec:conclusion}

In this paper, we have developed two novel Bayesian neural network-based multi-modal surrogate models. These models build upon existing work in multi-modal learning, Bayesian neural networks, and multi-fidelity surrogate modeling. We estimate the parameters of these models using variational Bayes estimation of these models' posterior distributions that leverages known and closed-form expressible posterior distributions in the last layer. We developed a novel technique to complete this estimation procedure in the presence of missing data by drawing the missing values from distributions that approximate the posterior predictive distribution of the missing data. Without this step, the missing data would make the last-layer posterior distributions have no known closed form. We demonstrate through simulation that these models show improved estimation for the main modality in some datasets when additional samples of auxiliary modalities are present. The canonical correlation between the auxiliary modality data, collectively, and the main modality or quantity of interest appears to be indicative to determine a priori whether these multi-modal models may lead to improved predictions compared to  uni-modal alternatives. While multi-fidelity surrogate models are often hierarchical, these multi-modal models consider alternate data sources equally and enable utilizing them in combination. A method that measures non-linear relationships in order to determine a priori the performance of these multi-modal models is left for future work.

%\julie{this is obscur to me, could you have tried to investigate that? are you saying that we missed an opportunnity?}With this work, it remains an open question when available multi-modal data sources are useful, i.e., aid surrogate model accuracy over the equivalent model without them. \textcite{toal_considerations_2015} and \textcite{andres-thio_characterising_2024} are notable works in this area for multi-fidelity Bayesian optimization using Gaussian process surrogate models. These works provide tools to evaluate multi-fidelity datasets to determine a priori whether multi-fidelity Bayesian optimization (co-Kriging) or standard Bayesian optimization should be used. However, the classification of helpful and harmful data sources defined by these works depend on the linear relationship between low and high fidelity data sources enforced in the co-Kriging model, and new classifications are needed with more flexible neural network surrogate models.

A multi-modal acquisition framework to complete the BO procedure using these multi-modal surrogate models is future work. Such a framework would guide both the input location, $\bm{x}^\ast$, and the modality at which to sample new data while optimizing the main objective function. It must be able to do this with the kind of non-linear, complex relationships between data modalities created by the BNN surrogate model, and account for the high-dimensional nature of some data modalities.

\section*{Acknowledgments}
\if0\blind{
This work was authored by the National Laboratory of the Rockies, operated by
Alliance for Energy Innovation, LLC, for the U.S. Department of Energy (DOE)
under Contract No. DE-AC36-08GO28308. Funding was provided by the Office
of Science, Office of Advanced Scientific Computing Research, Scientific Discovery
through Advanced Computing (SciDAC) program through the FASTMath Institute.
The views expressed in this article do not necessarily represent the views of the
DOE or the U.S. Government. The U.S. Government retains and the publisher, by
accepting the article for publication, acknowledges that the U.S. Government retains
a nonexclusive, paid-up, irrevocable, worldwide license to publish or reproduce the
published form of this work, or allow others to do so, for U.S. Government purposes.

This research was performed using computational resources sponsored by the Department of Energy’s Office of Critical Minerals and Energy Innovation and located at the National Laboratory of the Rockies.
\fi\if1\blind{\emph{Removed for anonymous copy.}}\fi

\section*{Disclosures}

The authors report there are no competing interests to declare.

\printbibliography

\appendix

\section{Simulated Datasets} \label{apx:simulated-datasets}

The \textbf{Branin} function \parencite{surjanovic_virtual_2013} is a function from $[-5,10]\times[0,15]$ to $\mathbb{R}$, consisting of a periodic component and a polynomial component. The ``high fidelity'' function is given by
\begin{equation}
    f(x_1,x_2) = a(x_2 - bx_1^2 + cx_1 - r)^2 + s(1-t)cos(x_1) + s,
\end{equation}
where $a=1$, $b=5.1/(4\pi^2)$, $c=5/\pi$, $r=6$, $s=10$, and $t=1/(8\pi)$. \textcite{toal_considerations_2015} define ``low fidelity'' versions of the Branin function through a parameter $A_1 \in [0,1]$:
\begin{equation}
    f'_{A_1}(x_1,x_2) = f(x_1,x_2) - (A_1 + 0.5)\cdot a(x_2 - bx_1^2 + cx_1 - r)^2,
\end{equation}
for the same values of $a$, $b$, $c$, and $r$, effectively adjusting the contribution of the polynomial component.

To create a training dataset, we generate data from the ``high fidelity'' function as the main modality, and from the ``low fidelity'' function for $A_1 \in \{0, 0.514, 1\}$ so that the low and high fidelity functions will have high correlation and low RMSE, low correlation and moderate RMSE, and high correlation and high RMSE, respectively. The main modality is sampled at the points $(x_1,x_2)\in\{-2, -0.5, 1, 4, 5.5, 7\}\times\{3, 4.5, 6, 9, 10.5, 12\}$, and the alternate modalities are sampled on the larger space $(x_1,x_2)\in\{-3.5, -2, -0.5, 1, 2.5, 4, 5.5, 7, 8.5\}\times\{1.5, 3, 4.5, 6, 7.5, 9, 10.5, 12, 13.5\}$. 

The \textbf{Paciorek} function \parencite{toal_considerations_2015, mainini_analytical_2022} is a function from $D$-dimensional space $\bm{x} \in [0.3, 1]^D$ to $\mathbb{R}$ consisting of periodic functions. The ``high fidelity'' version of the function is given by
\begin{equation}
    f(\bm{x}) = \sin\left(\prod_{i=1}^D x_i^{-1}\right),
\end{equation}
while the ``low fidelity'' versions of the function are parameterized by a value $A_2\in[0,1]$,
\begin{equation}
    f'_{A_2}(\bm{x}) = f(\bm{x}) - 9A^2_2\cos\left(\prod_{i=1}^D x_i^{-1}\right).
\end{equation}

To create a training dataset, we generate data from the ``high fidelity'' function as the main modality, and from the ``low fidelity'' function for the auxiliary modalities. We choose the four values $A_2 \in \{0.25, 0.5, 0.75, 1\}$ to create a range of auxiliary modalities from high correlation, low RMSE to the main modality, to low correlation, high RMSE. For the \textbf{Paciorek (high)} dataset, we instead choose $A_2 \in \{0.125, 0.25, 0.375, 0.5\}$, and for the \textbf{Paciorek (low)} dataset, we choose $A_2 \in \{0.625, 0.75, 0.875, 1.0\}$. This way, all versions of this dataset have the same number of auxiliary modalities and the only difference is the correlation between modalities. We choose $D=4$, and generate the main modality at the points $\bm{x} \in \{0.475, 0.5625, 0.7375, 0.825\}^4$, and the auxiliary modalities at those points, plus the points $\bm{x} \in \{0.3875, 0.51875, 0.65, 0.71825, 0.9125\}^4$.

The \textbf{Wind} data is comprised of an ERA5 dataset, a re-analysis dataset of meteorological variables produced by the European Centre for Medium-Range Weather Forecasts (ECMWF) and other institutions \parencite{hersbach_era5_2020}, and observational data from the Argonne National Laboratory tower measurements  \url{https://www.anl.gov/evs/atmos}. The dataset consists of hourly values of the following variables:
\begin{itemize}
    \item \texttt{obs\_Spd10m}: observed wind speed at 10 meters elevation
    \item \texttt{obs\_Spd60m}: observed wind speed at 60 meters elevation
    \item \texttt{era\_Spd10m}: modeled wind speed at 10 meters elevation
    \item \texttt{era\_Spd60m}: modeled wind speed at 60 meters elevation
    \item \texttt{era\_Spd100m}: modeled wind speed at 100 meters elevation
    \item \texttt{era\_u10m}, \texttt{era\_v10m}: zonal (west-east) and meridional (north-south) components of the modeled wind vector at 10 meters elevation
    \item \texttt{era\_u100m}, \texttt{era\_v100m}: zonal (west-east) and meridional (north-south) components of the modeled wind vector at 100 meters elevation
    \item \texttt{obs\_Dir10m}: observed wind direction (in degrees) at 10 meters elevation
    \item \texttt{obs\_Dir60m}: observed wind direction (in degrees) at 60 meters elevation 
    \item \texttt{era\_Dir10m}: modeled wind direction (in degrees) at 10 meters elevation
    \item \texttt{era\_Dir100m}: modeled wind direction (in degrees) at 100 meters elevation
\end{itemize}
To avoid wrap-around issues with the wind direction variables, we replace each of the four wind direction variables with their $\sin$ and $\cos$ to make 17 total variables.

We treat \texttt{obs\_Spd10m} as the main modality and the others as auxiliary modalities. All  modalities have physical complementarity. Model outputs and observations provide different and complementary representations of the same physical process due to their different sources of errors and uncertainties.  Meanwhile wind processes are defined through the entire atmospheric layers, creating correlation between wind at different vertical heights. Finally, wind speed and direction are coupled in intricate ways depending on many factors. The goal is to best surrogate \texttt{obs\_Spd10m} from the other modalities, knowing that in practice observations such as \texttt{obs\_Spd10m} are often harder to collect and get access to. We define the input, $x$, as the number of days past midnight, January 2, 2007. The data extend through December 31, 2021. For the purposes of evaluating our models, we chose only a small portion of the data from a range with no missing values. For fractional values of $x$ that fall between the hourly resolution of the dataset, we use cubic spline interpolation to produce intermediate sub-hourly values. We sample the main modality at the values $x \in \{k+\frac{i}{20}:i=0,\dots,20,k=2,4,6\}$ and the auxiliary modalities at the values $x \in \{k+\frac{i}{20}:i=0,\dots,20,k=1,2,3,4,5,6\}$.

For the \textbf{Wind (daily)} dataset, we restrict the input $x$ to be a whole number of days, and let each observation of a modality be a vector of all 24 hourly measurements from that day, rather than a scalar value for a specific time. We pull training and test data from the year 2019. All days in the months of February, April, June, July, September, and November are training data for the main modality, while all 365 days of the year are training data for the auxiliary modalities. Similar to the previous treatment of the wind data, we expect these modalities to complement each other because they are different and complementary representations of the same physical process. In this setting, however, their relationships must persist through the PCA decomposition of each modality. We expect this to happen because we keep enough PCA components to explain a large proportion of the variability in the data of each modality.

The \textbf{Time Series} data are simulated to provide an example with higher-dimensional modalities. The time series data are produced by a function $f:[0,1]\to\mathbb{R},$ defined by
\begin{equation}
    f(t;\alpha,\beta,\gamma,\delta) = t^\alpha + \beta\cos\left(2\pi \frac{t}{\gamma} + \delta\right),
\end{equation}
where $\alpha > 0$, $\beta > 0$, $\gamma > 0$, and $-\pi < \delta < \pi$.

\begin{figure}
    \centering
    \includegraphics[width=0.8\linewidth]{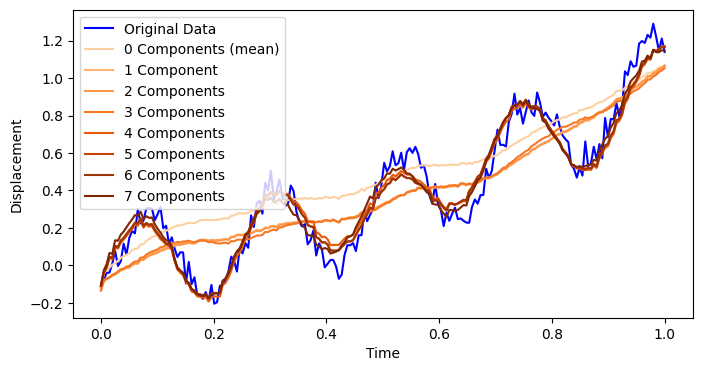}
    \caption{Example of the time series modality for the input parameters $\log\alpha = 0.5$, $\log\gamma = -1.5$, and $\mathrm{atanh}(\delta)=-1$, along with its reconstruction using 0 through 7 principle components. Before training, the collection of these training data would be used to fit a PCA decomposition, and the $k$ largest coefficients would be recorded to account for 95\% of the variability in the training data.}
    \label{fig:example-timeseries-data}
\end{figure}

For a chosen resolution, $N$, we generate a noisy time series of length $N$, $\bm{y}$, such that $y_i = f\left(\frac{i-1}{N-1};\alpha,0.25,\gamma,\delta\right) + \epsilon_i$, where $\epsilon_i \overset{\mathrm{i.i.d.}}{\sim} \mathrm{N}(0, 0.05^2)$. In addition to the fully observed noisy time series, we include three scalar modalities relating to the underlying (noiseless) function and acting as (partial) summary statistics:
\begin{itemize}
    \item Total Distance: $\int_0^1\left|\frac{\operatorname{d}f}{\operatorname{d}t}\right|\operatorname{d}t$
    \item Average Slope: $f(1;\alpha,\beta,\gamma,\delta) -f(0;\alpha,\beta,\gamma,\delta)$
    \item Maximum Location: $\arg\max_t f(t;\alpha,\beta,\gamma,\delta)$
\end{itemize}

For the ``Time Series'' dataset, we use $N=200$. We use a PCA-decomposed representation of the full-length $N=200$ noisy time series as the main modality, and all three scalar values as the auxiliary modalities. For ``Time Series (1D)'', we instead use Total Distance as the main modality and the PCA-decomposed time series and the other two scalar values as auxiliary modalities. We have one option with a scalar quantity of interest to better replicate downstream tasks such as optimization. Of all the scalar modalities, total distance is most affected by all of the input values. In both datasets, we observe the main modality at the points $(\log\alpha,\log\gamma,\mathrm{atanh}(\delta)) \in \{-1.5, -1, -0.5, 0.5, 1, 1.5\}^3$ and the auxiliary modalities at the points $(\log\alpha,\log\gamma,\mathrm{artanh}(\delta)) \in \{-2, -1.5, -1, -0.5, 0, 0.5, 1, 1.5, 2\}^3$. The number of PCA components is chosen to explain 95\% of the variability in the observed time series. Figure \ref{fig:example-timeseries-data} shows an example of the original noisy time series data along with the cumulative sum of the seven principal components that account for 95\% of the variation in the training data.

\end{document}